\documentclass{article} % For LaTeX2e
\usepackage[preprint]{colm2026_conference}

\usepackage{url}
\usepackage{lineno}
\usepackage{microtype}
\usepackage{graphicx}
\usepackage{booktabs}
\usepackage{multirow}
\usepackage{makecell}
\usepackage{booktabs}
\usepackage{amsmath}
\usepackage{amssymb}
\usepackage{mathtools}
\usepackage{amsthm}
\usepackage{listings}
\usepackage{xcolor}
\usepackage{hyperref}
\usepackage{subcaption}
\usepackage{xspace}

\newcommand{\llamathree}{Llama3.3-70B\xspace}
\newcommand{\llamafour}{Llama4-Maverick\xspace}
\newcommand{\qwensmall}{Qwen3-80B\xspace}
\newcommand{\qwenmedium}{Qwen3-235B\xspace}
\newcommand{\qwenlarge}{Qwen3-480B\xspace}
\newcommand{\gpt}{GPT-5.2\xspace}

\newcommand{\ecoparagraph}[1]{\vspace{0.1cm}\noindent\textbf{#1}}

\definecolor{darkblue}{rgb}{0, 0, 0.5}
\hypersetup{colorlinks=true, citecolor=darkblue, linkcolor=darkblue, urlcolor=darkblue}

\title{ReDAct: Uncertainty-Aware Deferral for LLM Agents}

\author{Dzianis Piatrashyn$^{1}$, Nikita Kotelevskii$^{1}$, Kirill Grishchenkov$^{2}$,\\
\textbf{Nikita Glazkov$^{3, 4}$, Ivan Nasonov$^{5}$, Ilya Makarov$^{4, 6}$, Timothy Baldwin$^{1}$, Preslav Nakov$^{1}$,}\\
\textbf{Roman Vashurin$^{1}$, Maxim Panov$^{1}$}\\[0.5em]
$^{1}$Mohamed bin Zayed University of Artificial Intelligence\\
$^{2}$Ocapital\\
$^{3}$National University of Science and Technology (NUST) MISIS\\
$^{4}$AXXX\\
$^{5}$Ivannikov Institute for System Programming of the Russian Academy of Sciences\\
$^{6}$Trusted AI Center, RAS\\
}

\newcommand{\xv}{\mathbf{x}}
\newcommand{\yv}{\mathbf{y}}

\newcommand{\SC}{\mathcal{S}}

\begin{document}

\ifcolmsubmission
\linenumbers
\fi

\maketitle

\begin{abstract}
Recently, LLM-based agents have become increasingly popular across many applications, including complex sequential decision-making problems. However, they inherit the tendency of LLMs to hallucinate, leading to incorrect decisions. In sequential settings, even a single mistake can irreversibly degrade the trajectory, making hallucinations an even bigger problem.
Although larger LLMs hallucinate less, they incur a significantly higher per-token cost.
In this paper, we address this tradeoff by proposing \textbf{ReDAct} \emph{(Reason-Defer-Act)}.
In ReDAct, an agent is equipped with two LLMs: a small, cheap model used by default, and a large, more reliable but expensive model.
When the predictive uncertainty of the small model exceeds a calibrated threshold, the decision is deferred to the large model.
We evaluate our approach in text-based embodied environments such as ALFWorld and MiniGrid and show that deferring only about 15\% of decisions to the large model can match the quality of using it exclusively, while significantly reducing inference costs.
\end{abstract}

% !TEX root = ../main.tex

\section{Introduction}
\label{sec:introduction}
  Large language models (LLMs; \citealp{brown2020language}) are widely used nowadays across a range of domains~\citep{bommasani2021opportunities,chowdhery2023palm}.
  Among many directions in which LLMs are deployed, one of the most recent and rapidly emerging is their use as autonomous agents~\citep{yao2022react,schick2023toolformer}.
  Here, the LLM autonomously plans and decides which actions to take at each step, given the observed context~\citep{yao2022react,ahn2022saycan,huang2022inner}.

  A particularly interesting instantiation of this paradigm appears in embodied and interactive environments.
  Text-based embodied benchmarks, such as ALFWorld~\citep{shridhar2021alfworldaligningtextembodied}, and grid-world environments, such as MiniGrid~\citep{MinigridMiniworld23}, place LLM agents in partially observed settings.
  Here, the agent's states change irreversibly, and errors in actions accumulate over the entire episode.
  A single wrong action, such as stepping onto a lava tile or picking up the wrong object, can immediately degrade the trajectory and ruin the entire episode.
  Unlike static text generation and other classical (e.g., QA) LLM setups, these environments require that agents make reliable decisions at every step, because errors compound rather than remain isolated~\citep{huang2022inner,wang2023voyager}.

  This motivates a mechanism that helps the agent recognize mistaken or unreliable actions. When such actions are detected, the agent can refuse to act and defer the decision-making to a more capable model.
  Classical selective prediction and learning-to-defer frameworks~\citep{geifman2019selectivenet,mozannar2020consistent} formalize this idea.
  In the LLM setting, the agent can abstain from making a decision when its small model is unsure and defer to a larger, more capable model.
  This formulation connects naturally to cost-aware model routing and cascading~\citep{chen2023frugalgpt,yue2023large,ong2024routellm}, where the goal is to use expensive large models only when they are likely to improve the outcome.
  However, existing routing methods treat model selection as a per-query problem, choosing a single model for the entire request. They do not consider the repeated setting of an agent that sequentially makes decisions, nor deferral within an interactive trajectory.

  Prior work on uncertainty quantification (UQ) has developed numerous methods that allow quantifying the reliability of LLMs' predictions~\citep{vashurin-etal-2025-benchmarking}. These include, \textit{inter alia}, semantic entropy~\citep{kuhn2023semantic}, information-theoretic approaches~\citep{fomicheva-etal-2020-unsupervised, malinin2021uncertaintyestimationautoregressivestructured}, verbalized confidence~\citep{tian-etal-2023-just} and others.
  However, these works normally evaluate uncertainty in single-turn language generation. Much less is known about the usefulness of uncertainty scores from a small LLM for reliable deferral to a large model in sequential environments.
  
  In this paper, we study this problem. We propose \textbf{ReDAct} \emph{(Reason-Defer-Act)}, a framework in which an agent is equipped with two LLMs: a small, cheap model and a large, more reliable but expensive one. When the small model's uncertainty is high, the agent defers the action decision to the large model.

  Our contributions are as follows.
  \begin{itemize}
    \item We propose ReDAct, a framework for LLM agents in sequential environments in which the base agent can defer individual action decisions from a small model to a large model based on predictive uncertainty (see Section~\ref{sec:method}).
    
    \item We measure the performance of multiple state-of-the-art UQ methods and show that action-level information-theoretic UQ measures provide the most informative deferral signal (see Section~\ref{sec:action_vs_reasoning_level_uncertainty}).

    \item We evaluate our approach on ALFWorld and MiniGrid datasets using frontier-size LLMs (70B to 480B) and demonstrate that deferring only about 15\% of decisions to the larger model can match its full performance, at a fraction of the inference cost (see Section~\ref{sec:experiments}).
  \end{itemize}

% !TEX root = ../main.tex

\section{Related Work}
\label{sec:related_work}
  Our work is placed at the intersection of several topics, namely uncertainty quantification, prediction deferral, and the application of agents in embodied environments. Let us review these areas one by one.

\ecoparagraph{Uncertainty Quantification.}
  The area of predictive uncertainty quantification has been developing rapidly in recent years, given the increased scope of problems in which predictive models are applied~\citep{kendall2017uncertainties,gal2016dropout,hullermeier2021aleatoric,kotelevskii2025from}.
  As these problems involve high-risk applications, such as autonomous driving~\citep{bojarski2016end} and medical diagnostics~\citep{kompa2021second}, for reliable use in real scenarios, one needs to be careful with model predictions and detect unreliable ones.
  The problem of predictive uncertainty quantification has been extensively studied across different domains~\citep{gawlikowski2023survey,abdar2021review}, and interest in uncertainty for LLMs has grown particularly rapidly in recent years.
  There are several key groups of methods for quantifying uncertainty in LLM predictions. We will emphasize three of them.

  The first is an information-theoretic group, which works directly with the probabilities of generated tokens~\citep{fomicheva-etal-2020-unsupervised, malinin2021uncertaintyestimationautoregressivestructured}. As the methods in this class require token or sequence probabilities, the model must be white-box in this sense.
  Another group works with the semantics of generated outputs~\citep{lingenerating, kuhn2023semantic}. These methods typically require multiple sequence generations from the LLM, grouping them into semantically close clusters using Natural Language Inference (NLI), and computing a measure of ``variance'', e.g., entropy, over them. Hence, these methods may be computationally more demanding.
  Recently, approaches that bridge these two classes of methods have emerged~\citep{vashurin2025uncertainty}.
  The third group we want to emphasize includes the so-called verbalized methods. These methods prompt the LLM to output an explicit ``I am not sure'' token and estimate its probability~\citep{kadavath2022language}, or ask the model to output its confidence directly~\cite{tian-etal-2023-just}.

  There is no established leader among these methods, as different combinations of models, datasets, and problems can yield different rankings.

\ecoparagraph{Prediction deferral.}
  The option of abstaining from a prediction has been considered in the literature for a general statistical problem setting~\citep{chow2003optimum,geifman2019selectivenet}.
  It was also considered in different domains, such as images~\citep{geifman2019selectivenet} and text~\citep{malinin2021uncertaintyestimationautoregressivestructured}.
  Deferral to a more advanced model was studied in both the image domain~\citep{kotelevskiidirichlet} and the text domain~\citep{mozannar2020consistent,verma2022calibrated}.
  For LLMs, the switch from a small model to a larger, more capable one was considered in~\citep{chen2023frugalgpt,yue2023large,ong2024routellm}. In particular, \citet{ong2024routellm} trained a supervised router to select between models. However, these works studied only single-step setups, e.g., QA, without a sequential agentic problem formulation.

  A notable exception is~\citep{ren2023robots}, which defers the decision to the oracle in steps where the predictive uncertainty is high. Although important, this work considers only conformal prediction, which requires scoring each possible action using an LLM. This can be cost-prohibitive in settings with a large number of possible actions at each step. Moreover, it considers only deferral to a human oracle, while we keep the system fully autonomous.

\ecoparagraph{Agents in embodied systems.}
  Several works consider agents operating in embodied environments. 
  ALFWorld~\citep{shridhar2021alfworldaligningtextembodied} is a text-based counterpart of the ALFRED benchmark~\citep{shridhar2020alfred}.
  It consists of long compositional household tasks with irreversible state changes. 
  MiniGrid~\citep{MinigridMiniworld23} consists of grid-world environments that test navigation and exploration under partial observability.
  WebShop~\citep{yao2022webshop} simulates an online shopping task that requires multi-step search, comparison, and purchase decisions.
  LLM-based agents have been applied to these environments with growing success~\citep{yao2022react,wang2023voyager,shinn2023reflexion}.

  However, in these settings, uncertainty-aware deferral between small and large LLMs within an agent has not been considered.

% !TEX root = ../main.tex

  \begin{figure}[t]
    \centering
    \includegraphics[width=\linewidth]{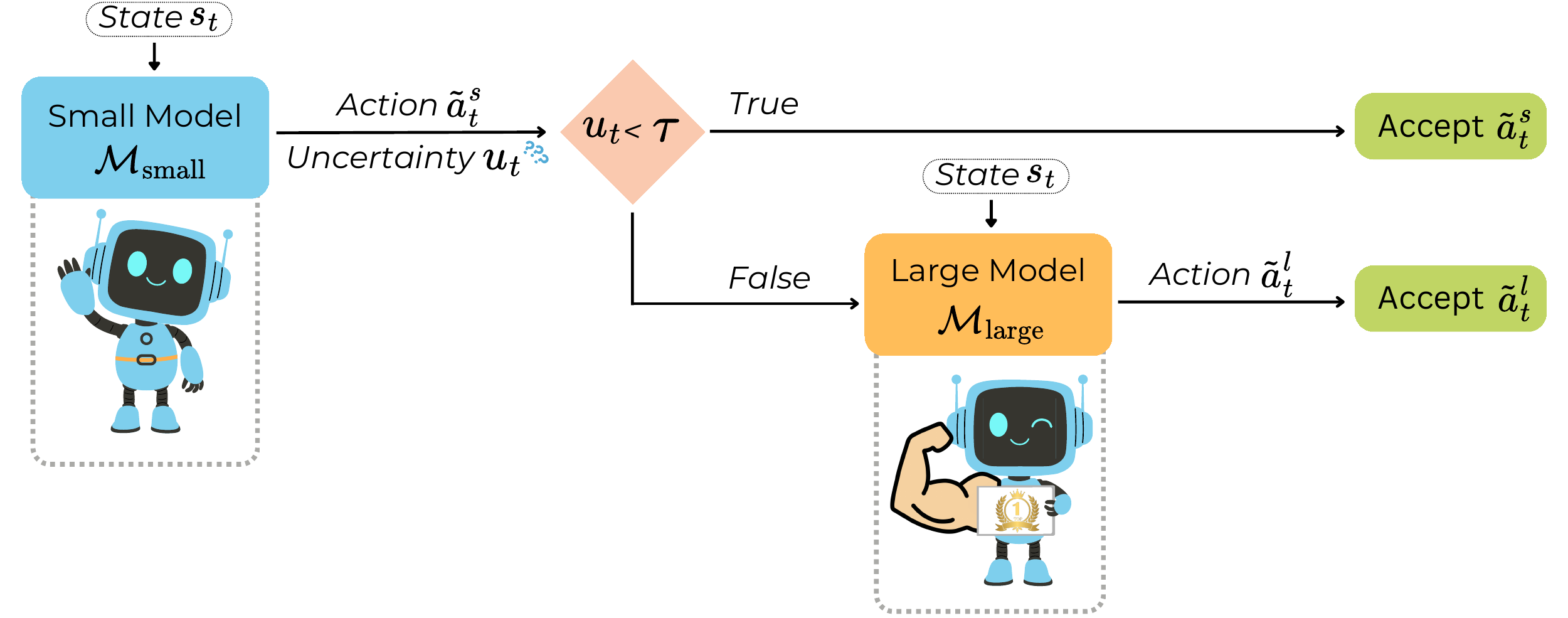}
    \caption{Overview of the proposed ReDAct framework. The agent uses the small model to propose an action \( \tilde{a}^s_t \) and estimates its uncertainty \( u_t \). If \( u_t \) exceeds the threshold \( \tau \), the step is deferred to the large model, which proposes \( \tilde{a}^l_t \) as the final decision.}
    \label{fig:deferral_scheme}
  \end{figure}

\section{Background}
\label{sec:background}
  We start this section by formalizing our setup and introducing useful notation.

\ecoparagraph{Problem Statement.}
  We consider an agent placed in an environment that it can interact with.
  The agent is equipped with two LLMs: a small, cheap model \( \mathcal{M}_{\text{small}} \) used by default, and a large, more reliable but expensive model \( \mathcal{M}_{\text{large}} \).
  The goal of the agent is to complete a task (e.g., find an item in a maze), described in natural language in the input prompt, and do so effectively (e.g., in as few steps as possible).
  Each such interaction with the environment forms an \textit{episode}, which can end either with success or failure.

  The episode consists of discrete steps \( t \in \{1, 2, \ldots, T\} \).
  At each step \( t \), the agent observes the current state of the environment \( s_t \in \SC \).
  Then, using the small model, the agent generates a proposal action \( \tilde{a}^s_t  \), and estimates the uncertainty score \( u_t \) associated with this proposal.
  If \( u_t \) is below a selected threshold \( \tau \) (see Section~\ref{sec:method}), the proposed action is accepted, \( a_t \coloneqq \tilde{a}^s_t \).
  Otherwise, the proposed action is rejected, and the decision is deferred to the large model, which generates its own proposal \( \tilde{a}^l_t \), accepted unconditionally: \( a_t \coloneqq \tilde{a}^l_t \).
  In both cases, the accepted action \( a_t \) is applied to the environment, transitioning the agent to a new state \( s_{t+1} \).
  A schematic illustration of our approach is provided in Figure~\ref{fig:deferral_scheme}.
  The episode ends when the task is completed (success) or when the agent reaches the maximum number of steps \( T \) (failure).

  If cost were not a concern, a rational actor would use the large model at every step.
  In practice, however, a tradeoff between the two models is desirable.

  We propose balancing this tradeoff using uncertainty quantification, where the uncertainty score associated with a proposed action is compared to a threshold to decide when deferral is needed.

\ecoparagraph{Uncertainty Quantification.}
  Let us denote by \( \yv = [y_1, y_2, \ldots, y_L] \) a sequence, generated by an LLM for an input prompt \( \xv = [x_1, x_2, \ldots, x_n] \).
  One can compute the probability of the output sequence \( \yv \) as:
  \begin{equation}
    p(\yv \mid \xv, \theta) = \prod_{i=1}^{L} p(y_i \mid \xv, \yv_{<i}, \theta),
  \end{equation}
  where \( \theta \) are the model parameters, and \( \yv_{<i} = (y_1, \ldots, y_{i-1}) \).

  If these token-level probabilities are available, one can compute information-theoretic measures of predictive uncertainty. 
  In this work, we consider three measures that use the small model's output probabilities directly, specifically Sequence Probability (SP), Perplexity (PPL), and Mean Token Entropy (MTE).
  All three aggregate token-level probabilities into a single scalar uncertainty score for the generated sequence. Their formal definitions are given in Appendix~\ref{appendix:uq_details}.

  These methods require a token probability distribution but, unlike methods based on semantic similarity, do not require multiple generations or external NLI models.
  Hence, these methods are computationally cheaper and well-suited for step-level deferral decisions in a sequential setting. As we will see in Section~\ref{sec:experiments}, they also demonstrate the most promising results.
  Note that in general, one can consider multimodal environments (e.g., visual-language ones). In this scenario, one may consider visual-based uncertainty quantification measures.

  Higher uncertainty scores result in lower confidence in the generated action. When \( u_t > \tau \), the small model's proposal \( \tilde{a}^s_t \) is considered unreliable, and the step is deferred to the large model.

\ecoparagraph{ReAct Framework.}
  In our setting, we start with the ReAct framework~\citep{yao2022react}, which decouples reasoning and action selection into two separate LLM calls, and extend it with deferral capabilities.
  At each step, the agent first generates a reasoning process that analyzes the current position, then uses this reasoning to select the next action.
  Separating reasoning and action allows us to incorporate deferral naturally. Specifically, we compute the uncertainty score at the action-selection stage, after the small model's reasoning is complete.
  If the uncertainty is high, the action call is deferred to the large model. The large model makes its own reasoning from scratch and returns the action \( \tilde{a}^l_t \), which the agent accepts.
  Using the prompt from Figure~\ref{fig:alfworld_reasoning_prompt}, we obtain the reasoning, which we then use as input for the action-selection prompt presented in Figure~\ref{fig:alfworld_action_prompt} (see Appendix~\ref{appendix:prompts}).

% !TEX root = ../main.tex

\begin{table}[!t]
    \centering
    \begin{tabular}{l|cc|cc}
        \toprule
        & \multicolumn{2}{c|}{\textbf{Action}} & \multicolumn{2}{c}{\textbf{Reasoning}} \\
        \textbf{UQ metric} & \textbf{PRR} & \textbf{ROC-AUC} & \textbf{PRR} & \textbf{ROC-AUC} \\
        \midrule
        Mean Token Entropy & \textbf{0.424} & \textbf{0.710} & 0.168 & 0.596 \\
        Perplexity & 0.399 & 0.684 & 0.193 & 0.616 \\
        Sequence Probability & 0.417 & 0.704 & \textbf{0.279} & \textbf{0.682} \\
        \bottomrule
    \end{tabular}
    \caption{Performance comparison of uncertainty metrics estimated over action selection and reasoning trace.}
    \label{tab:uq_comparison}
\end{table}

\section{Method}
\label{sec:method}
  In this section, we describe the ReDAct framework in detail.
  The method involves two key choices.
  First, we need to determine the stage at which uncertainty is computed and which class of uncertainty measures to use.
  Second, we need to select the deferral threshold. We describe both below.

\subsection{Action vs Reasoning-Level Uncertainty}
\label{sec:action_vs_reasoning_level_uncertainty}
  As we follow the ReAct framework, we can defer to the large model when the uncertainty in the reasoning stage or the action selection step is significant.
  To determine which stage provides a more reliable deferral signal, we conduct an auxiliary experiment on the ALFWorld dataset.
  Specifically, we simulate 100 trajectories with the ReAct framework using the \qwensmall model, resulting in a total of 2411 steps.

  The environment does not provide ground-truth actions for intermediate steps. To recover correctness labels, we use \gpt as an LLM-as-a-judge~\citep{zheng2023judging} to annotate each step as either correct (meaningful exploration or appropriate action selection) or incorrect (e.g., cyclic behavior or repeatedly visiting the same room). 
  After this labelling procedure, we obtain 1555 steps labelled as correct and 856 as incorrect. We then compute UQ metrics for this labeled dataset and evaluate them via PRR and ROC-AUC.

\ecoparagraph{Results.}
  We report the results for reasoning-level and action-level UQ in Table~\ref{tab:uq_comparison} (see more results in Appendix~\ref{appendix:full_uq_results}).
  For UQ methods estimated at the action selection step, we observe high PRR and ROC-AUC values, especially for Perplexity, Sequence Probability, and Mean Token Entropy.
  Apart from superior performance, these three methods do not require repeated model inference, making them ideal for both UQ performance and inference costs.
  In contrast, UQ methods calculated at the reasoning step show poor discriminative power. This result is not surprising, as uncertainty quantification over free-form reasoning is known to be a difficult problem~\citep{ni2025efficient}.
  Based on this finding, ReDAct focuses exclusively on action-level UQ metrics.

\subsection{Threshold Selection}
\label{sec:threshold}
  A central component of ReDAct is the selection of the deferral threshold \( \tau \). Specifically, if the uncertainty at the action stage exceeds \(\tau \), the decision is deferred to the large model.
  We calibrate the threshold using a set of \( N_{\text{cal}} \) calibration episodes (the same episodes for which we reported results in Table~\ref{tab:uq_comparison}).
  The calibration procedure is as follows. We fix a target average number of large model calls per episode, denoted \( K \).
  For each UQ measure (SP, MTE, PPL), every uncertainty value observed across all steps and episodes is treated as a candidate threshold.
  We then select the candidate that yields an average number of large model invocations per episode closest to \( K \) on the calibration set.
  Note that the number of large model calls is monotonically non-increasing in the threshold, so this selection is well-defined and can also be performed effectively via binary search.
  For the random baseline, we set the deferral probability uniformly at each step to \( K / \bar{L} \), where \( \bar{L} \) is the average number of steps per episode on the calibration set.

\ecoparagraph{Limitations.}
  Although we avoid using the expensive large model during calibration, this introduces a distribution shift in the uncertainty scores.
  Specifically, in the procedure described above, we select the threshold based solely on the distribution of uncertainty values produced by the small model.

  However, once the large model is invoked during evaluation, it changes the distribution of uncertainty scores at subsequent steps. This results in a slight discrepancy between the expected and actual number of large model requests.
  As our experimental results show, this effect is not severe in practice. Furthermore, it could be eliminated by performing calibration during a warm-up run with enabled deferral. The initial deferral threshold would be selected and continuously updated until the desired average number of deferrals is achieved.

% !TEX root = ../main.tex

\section{Experiments}
\label{sec:experiments}
  In this section, we evaluate ReDAct on text-based interactive environments.
  We consider both task performance and the economic benefits of uncertainty-based deferral.
  Note that while our results are focused on text-based environments, our methodology is not limited to them and can be extended to other modalities, e.g., visual-textual.

\subsection{Experimental Setup} 
  We start by describing in detail our experimental setup, including the datasets, models, metrics, and baselines used for evaluation.

\ecoparagraph{Datasets.} 
  In our experiments, we consider two text-based embodied environments, specifically ALFWorld~\citep{shridhar2021alfworldaligningtextembodied} and MiniGrid~\citep{MinigridMiniworld23}.

  \begin{itemize}
    \item For \textbf{ALFWorld}, we considered 400 episodes, with the maximum number of steps per episode capped at 50. For more details see Appendix~\ref{appendix:alfworld_additional_details}.
    
    \item For \textbf{MiniGrid}, we considered 200 episodes, with the maximum number of steps per episode capped at 50. Note that in the main body of the paper, we use a full-view setting for MiniGrid. The reason is that the majority of models we considered (including large models) failed to produce reasonable results in a partial-view setup. For more details and partial-view results, see Appendix~\ref{appendix:minigrid_additional_details}.
  \end{itemize}

\ecoparagraph{Models and Prices.}
  As described in Section~\ref{sec:background}, the agent is equipped with two LLMs.
  The first is a small, inexpensive model that can even be run locally, without paying high per-token usage fees. However, it may be more prone to hallucinations than its expensive counterpart. As small models, we consider \qwensmall, \llamathree, and \llamafour.

  The second is a large, more reliable model with much higher per-token usage costs. As large models, we consider \gpt, \qwenmedium, and \qwenlarge.

  To infer these models (except for \gpt), we use the Together AI service.\footnote{\url{https://www.together.ai/}} More details on the models we use are provided in Appendix~\ref{appendix:models_additional_details}.

  We also provide a comparative table with input/output token prices for different models in Table~\ref{tab:price_list}.

\ecoparagraph{Metrics.}
  As described in Section~\ref{sec:background}, each episode has a binary outcome: success (1) or failure (0).
  The episode is considered successful if the agent achieves the goal within 50 steps. Otherwise, it is considered a failure.
  Therefore, one natural measure of an agent's performance is the \emph{success rate}, computed as the ratio of successful episodes to the total number of episodes:
  \begin{equation*}
    \text{Success Rate} = \frac{\#\text{Successful episodes}}{\#\text{Total episodes}}.
  \end{equation*}

  Another important dimension is cost. 
  As shown in Table~\ref{tab:price_list}, large models are significantly more expensive than small models.
  Therefore, we also report the overall inference expenses alongside downstream performance. When running only the large or small model without deferral, the cost reduces to that of the corresponding model alone.
  For ReDAct, the reported cost is the cumulative cost of the whole agent (small and large models).
  Ideally, one would like to achieve a success rate comparable to that of the large model alone, but at a much lower cost. ReDAct is designed to achieve this tradeoff.

\ecoparagraph{Baselines for deferral.} 
  As discussed in Section~\ref{sec:method}, we focus on action-level uncertainty measures for deferral. Among those, information-theoretic measures (MTE, SP, and PPL) showed the best discriminative performance and are therefore used as the main deferral indicator in ReDAct.
  We calibrate the deferral threshold on \( N_{\text{cal}} = 100 \) episodes with a target of \( K = 5 \) large model calls per episode on average.
  As an additional baseline, we consider \emph{random} large model invocation (see Section~\ref{sec:method}). Below, we report results for all of these choices.

\begin{table*}[t]
\centering
\small
\begin{tabular}{ll|cccc}
    \toprule
    \textbf{Base} & \textbf{Setting} & MTE & SP & PPL & Random \\
    \midrule
    
    \multirow{3}{*}{\qwensmall}
        & Base         & \multicolumn{4}{c}{0.683 $\pm$ 0.025} \\
        & + \gpt       & 0.788 $\pm$ 0.020 & \underline{0.798 $\pm$ 0.025} & \textbf{0.808 $\pm$ 0.011} & 0.770 $\pm$ 0.026 \\
        & + \qwenlarge & 0.743 $\pm$ 0.020 & \underline{0.763 $\pm$ 0.014} & \textbf{0.790 $\pm$ 0.023} & 0.760 $\pm$ 0.029 \\
    
    \midrule
    
    \multirow{3}{*}{\llamathree}
        & Base         & \multicolumn{4}{c}{0.635 $\pm$ 0.018} \\
        & + \gpt       & \underline{0.768 $\pm$ 0.017} & 0.763 $\pm$ 0.025 & \textbf{0.800 $\pm$ 0.028} & 0.750 $\pm$ 0.025 \\
        & + \qwenlarge & 0.755 $\pm$ 0.020 & \underline{0.760 $\pm$ 0.028} & \textbf{0.773 $\pm$ 0.017} & 0.690 $\pm$ 0.026 \\
    
    \midrule
    \gpt       & Base & \multicolumn{4}{c}{0.783 $\pm$ 0.019} \\
    \qwenlarge & Base & \multicolumn{4}{c}{0.793 $\pm$ 0.024} \\
    
    \bottomrule
\end{tabular}
\caption{Success rate on ALFWorld with bootstrapped standard deviation. For the small model without deferral, all methods coincide.}
\label{tab:alfworld_main_results}
\end{table*}

  \begin{table*}[t]
\centering
\small
\begin{tabular}{ll|cccc}
    \toprule
    \textbf{Base} & \textbf{Setting} & MTE & SP & PPL & Random \\
    \midrule
    
    \multirow{3}{*}{\qwensmall}
        & Base         & \multicolumn{4}{c}{0.855 $\pm$ 0.024 } \\
        & + \gpt    & 0.930 $\pm$ 0.015 & \underline{0.950 $\pm$ 0.013} & \textbf{0.950 $\pm$ 0.013} & 0.930 $\pm$ 0.019 \\
        & + \qwenmedium & \underline{0.930 $\pm$ 0.013} & 0.915 $\pm$ 0.017 & \textbf{0.940 $\pm$ 0.020} & 0.900 $\pm$ 0.012 \\
    
    \midrule
    
    \multirow{3}{*}{Llama4}
        & Base         & \multicolumn{4}{c}{0.220 $\pm$ 0.024} \\
        & + \gpt    & 0.370 $\pm$ 0.050 & 0.325 $\pm$ 0.030 & \textbf{0.420 $\pm$ 0.023} & \underline{0.380 $\pm$ 0.039} \\
        & + \qwenmedium & 0.330 $\pm$ 0.021 & \underline{0.355 $\pm$ 0.047} & \textbf{0.370 $\pm$ 0.031} & 0.295 $\pm$ 0.027 \\
    
    \midrule
    \gpt & Base    & \multicolumn{4}{c}{0.990 $\pm$ 0.007} \\
    \qwenmedium & Base & \multicolumn{4}{c}{0.925 $\pm$ 0.023} \\
    
    \bottomrule
\end{tabular}
\caption{Success rate on MiniGrid full-view with bootstrapped standard deviation. For the small model without deferral, all methods coincide.}
\label{tab:minigrid_main_results}
\end{table*}

\subsection{Results}
  Tables~\ref{tab:alfworld_main_results} and~\ref{tab:minigrid_main_results} present the main results of applying ReDAct to ALFWorld and MiniGrid, respectively.
  Every combination of small model, large model, and dataset shows that deferring to the large model guided by the UQ signal from the small model is superior to random deferral.
  Almost any choice of UQ measure improves the small model's success rate, but the best results are achieved with Perplexity~(PPL).
  Most notably on ALFWorld, PPL-based deferral achieves the full performance of the large model while invoking it for only \( {\sim} 15\% \) of the total number of steps.

\begin{figure}[t!]
    \centering
    \includegraphics[trim={0.cm 0.cm 0.cm 0.cm},clip,width=0.6\linewidth]{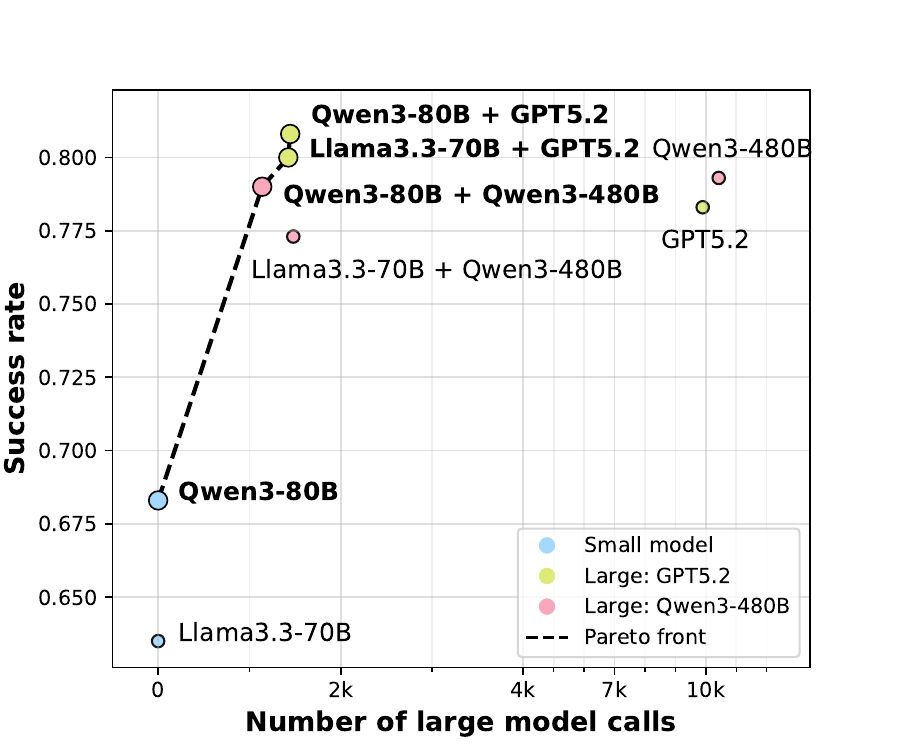}
    \caption{\textbf{Pareto front of success rate vs. large model calls on ALFWorld with perplexity-based deferral.} ReDAct enables small models to approach large model performance while using only ~15\% of large model calls.}
    \label{fig:pareto_alfworld_calls}
  \end{figure}

  Additionally, we show that UQ-guided deferral achieves Pareto-optimality both with respect to raw number of large model calls (Figures~\ref{fig:pareto_alfworld_calls} and~\ref{fig:pareto_minigrid_calls}) and to the actual inference costs (Figures~\ref{fig:pareto_alfworld_cost_combined} and~\ref{fig:pareto_minigrid_cost_combined}). This further strengthens our claim that cost-aware model routing is useful, and UQ guidance provides a valid signal for it.

  %%%%%%%

  \begin{figure}[b!]
    \centering
    \begin{subfigure}[t]{\linewidth}
    \includegraphics[trim={0cm 0cm 0cm 0cm},clip,width=0.9\linewidth]{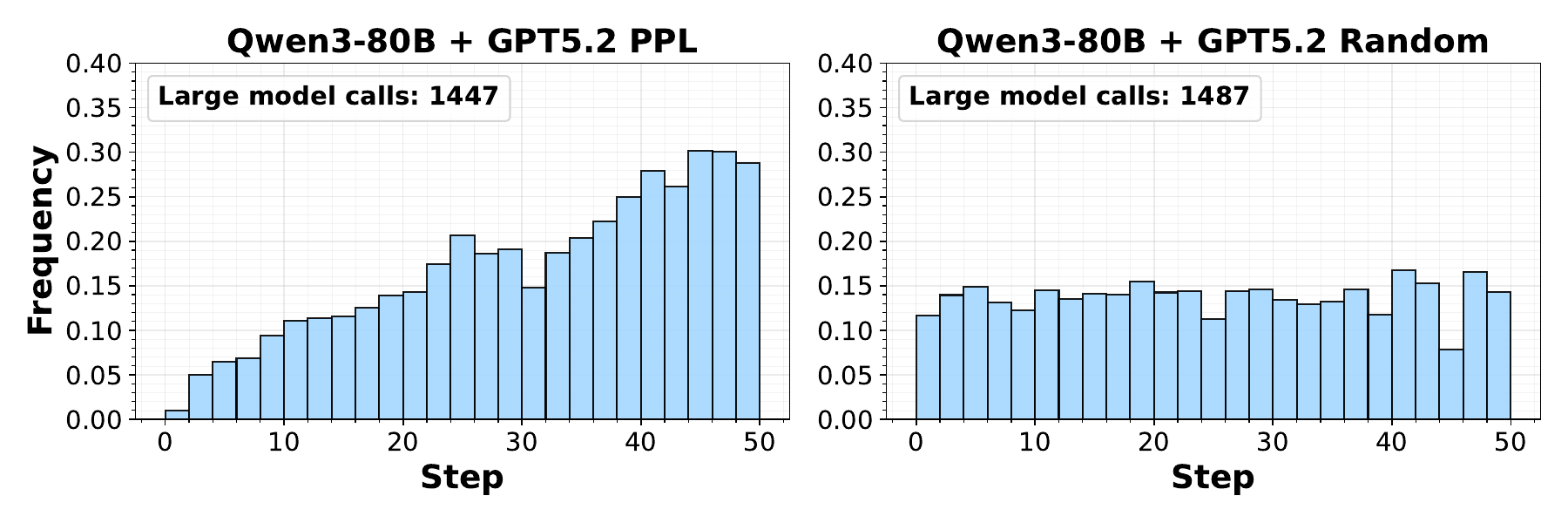}
    \end{subfigure}
    
    \begin{subfigure}[t]{\linewidth}
    \includegraphics[trim={0cm 0cm 0cm 0cm},clip,width=0.9\linewidth]{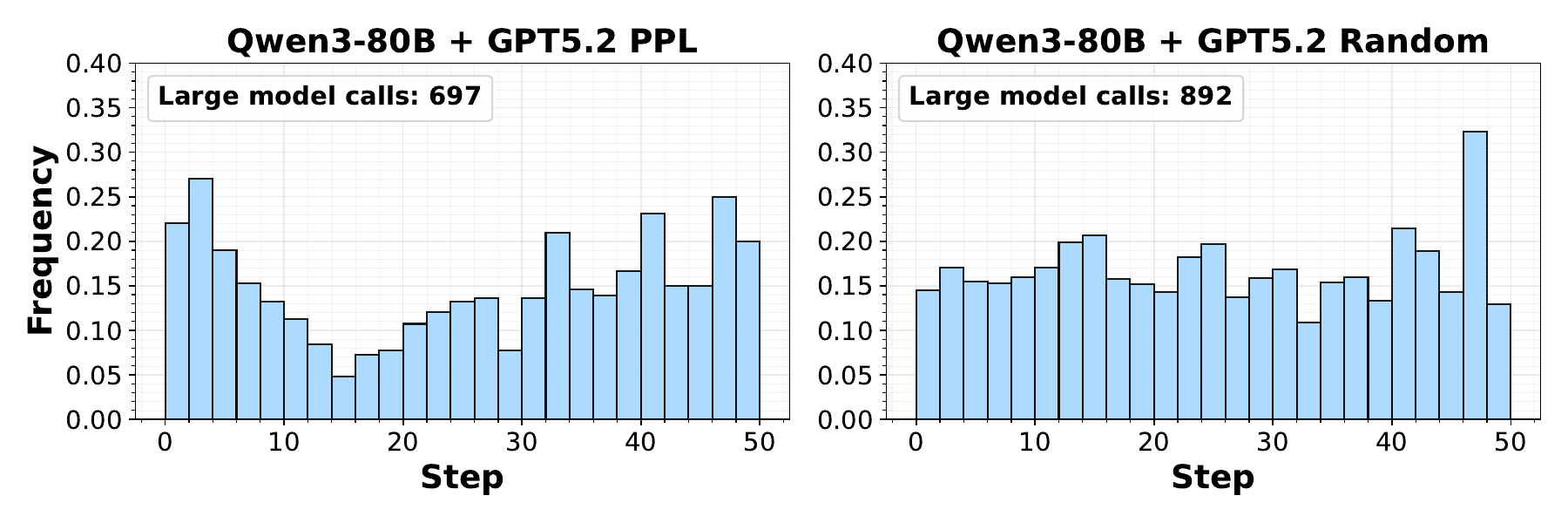}
    \end{subfigure}

    \caption{\textbf{\qwensmall + \gpt.} Large model invocation frequency by step in ALFWorld (top row) and MiniGrid (bottom row). At each step, the frequency is defined as the number of large model calls divided by the number of episodes that reached that step.}
  \label{fig:num_oracle_calls_qwen_gpt_shrinked}
  \end{figure}

\ecoparagraph{Distribution of calls.}
  Figure~\ref{fig:num_oracle_calls_qwen_gpt_shrinked} provides further evidence of the distinct nature of UQ-based deferral compared to the random approach.
  The figure shows that the relative frequency of large model calls per step for random deferral results in an almost perfectly uniform distribution, as expected. 
  For UQ-guided deferral distinct behavior is observed between datasets. For ALFWorld (see Figure~\ref{fig:num_oracle_calls_qwen_gpt_shrinked}, top row) the distribution is heavily skewed to the right. This explains why UQ-based deferral makes inference slightly more expensive on ALFWorld than random baseline, as shown in Table~\ref{tab:alfworld_costs}.
  During unsuccessful episodes, the small model becomes more uncertain as it digs itself deeper into a sub-optimal trajectory, and the likelihood of its uncertainty exceeding the deferral threshold increases.
  Indeed, for sequential agentic tasks, later steps have strictly longer prompts as they contain the full history of previous steps. Given the same average number of large model calls for random and UQ-based approaches, this results in the average large model call having a higher count of input tokens, thus making it marginally more expensive.
  This intuition is further confirmed by Tables~\ref{tab:alfworld_detailed_tokens_qwen_gpt},~\ref{tab:alfworld_detailed_tokens_qwen_qwen},~\ref{tab:alfworld_detailed_tokens_llama_gpt} and~\ref{tab:alfworld_detailed_tokens_llama_qwen} (see Appendix~\ref{appendix:additional_experimental_results}).
  The total count of input tokens is much higher for UQ-based setups, but the final cost is only slightly increased due to the low per-token cost of inputs.

  For MiniGrid (see Figure~\ref{fig:num_oracle_calls_qwen_gpt_shrinked}, bottom row), the distribution of large model calls has a distinct bimodal characteristic. This implies severe dependency of the complexity of the task on the initial position of the agent. An unfavorable start leads to an early deferral, producing the first mode. The second mode can be explained by the same argument, as the one used for ALFWorld.

  The distributions for remaining combinations of models can be found in Appendix~\ref{appendix:additional_experimental_results}.

  We report additional experimental results in Appendix~\ref{appendix:additional_experimental_results}, including number of large model calls, detailed token counts, and average number of steps per episode across all experimental setups.

\ecoparagraph{Cost-effectiveness.}
  To demonstrate the cost-effectiveness of ReDAct, we report actual inference costs in USD incurred by each evaluation setup on the ALFWorld benchmark in Table~\ref{tab:alfworld_costs}.
  It is evident that using the large model only when it matters results in real cost reduction without losing performance.
  Compared to the cost of using the large model exclusively, PPL-based deferral increases inference costs only modestly relative to running the small model alone.

  \begin{table*}[t]
\centering
\small
\begin{tabular}{ll|cccc}
    \toprule
    \textbf{Base} & \textbf{Setting} & MTE & SP & PPL & Random \\
    \midrule
    
    \multirow{2}{*}{\qwensmall}
        & Base         & \multicolumn{4}{c}{9.22} \\
        & + \gpt    & 15.43 & 16.39 & 16.25  & 14.15 \\
        & + \qwenlarge & 16.3 & 16.15 & 15.39 & 13.84 \\
    
    \midrule
    
    \multirow{2}{*}{\llamathree}
        & Base         & \multicolumn{4}{c}{29.24} \\
        & + \gpt    & 31.56 & 30.72 & 30.76 & 29.27 \\
        & + \qwenlarge & 33.05 & 32.18 & 33.21 & 31.82 \\
    
    \midrule
    \gpt & Base & \multicolumn{4}{c}{45.21} \\
    \qwenlarge & Base & \multicolumn{4}{c}{52.44} \\
    
    \bottomrule
\end{tabular}
\caption{Costs for inference per model in \$ for ALFWorld. Total is \$512}
\label{tab:alfworld_costs}
\end{table*}

% !TEX root = ../main.tex

\section{Conclusion}
\label{sec:conclusion}
  In this paper, we propose ReDAct, an uncertainty-aware deferral framework for LLM agents operating in sequential environments. The agent is equipped with two LLMs and defers individual action decisions from the small model to the large model when the small model's uncertainty is high.
  We evaluated our approach on text-based embodied environments and demonstrated that deferring only about 15\% of decisions can match, and sometimes surpass, the performance of the large model alone at a fraction of the cost.
  We also found that action-level uncertainty provides the most effective deferral signal, with information-theoretic measures such as Perplexity performing particularly well.
  We believe that adaptive switching between models within an agent is a promising direction for cost-efficient deployment, and that ReDAct offers a simple yet effective approach.

\section{Limitations}
  While ReDAct significantly improves the performance of small models on agentic tasks at a fraction of the cost of using large models exclusively, our study has certain limitations.
  First, ReDAct relies on information-theoretic UQ methods, which require access to token-level probabilities. In practice, however, most models of the size considered here expose logprobs through inference provider APIs.
  Second, we consider only models with more than 70B parameters. This choice is dictated by task complexity: smaller models perform poorly on agentic tasks without specialized fine-tuning and are thus outside the scope of this work.

% !TEX root = ../main.tex

\section*{Acknowledgements}
\label{sec:acknowledgements}

Part of the computational resources for our experiments were provided by Together AI.

\newpage

\bibliography{references}
\bibliographystyle{colm2026_conference}

%%%%%%%%%%%%%%%%%%%%%%%%%%%%%%%%%%%%%%%%%%%%%%%%%%%%%%%%%%%%%%%%%%%%%%%%%%%%%%%
%%%%%%%%%%%%%%%%%%%%%%%%%%%%%%%%%%%%%%%%%%%%%%%%%%%%%%%%%%%%%%%%%%%%%%%%%%%%%%%
% APPENDIX
%%%%%%%%%%%%%%%%%%%%%%%%%%%%%%%%%%%%%%%%%%%%%%%%%%%%%%%%%%%%%%%%%%%%%%%%%%%%%%%
%%%%%%%%%%%%%%%%%%%%%%%%%%%%%%%%%%%%%%%%%%%%%%%%%%%%%%%%%%%%%%%%%%%%%%%%%%%%%%%
\newpage

\appendix
% !TEX root = ../main.tex

\section{Measures of Uncertainty}
\label{appendix:uq_details}

We are using the following UQ methods:

  \begin{itemize}
    \item Sequence Probability (SP):
      \begin{equation}
        \text{U}_{\text{SP}} = -\sum_{i=1}^L \log p(y_i \mid \xv, \yv_{<i}, \theta).
      \end{equation}

    \item Perplexity (PPL):
      \begin{equation}
        \text{U}_{\text{PPL}} = -\frac{1}{L}\sum_{i=1}^L \log p(y_i \mid \xv, \yv_{<i}, \theta).
      \end{equation}

    \item Mean Token Entropy (MTE):
      \begin{equation}
        \text{U}_{\text{MTE}} = \frac{1}{L}\sum_{i=1}^L H(y_i \mid \xv, \yv_{<i}, \theta),
      \end{equation}
        where $H(y_i \mid \xv, \yv_{<i}, \theta)$ is the entropy of the token distribution $p(y_i \mid \xv, \yv_{<i}, \theta)$.
  \end{itemize}

\section {Additional Details on Evaluation Setup}
    \subsection{ALFWorld}
    \label{appendix:alfworld_additional_details}
    For ALFWorld, we report results on \textit{AlfredTWEnv} type with all available task types: Pick \& Place, Examine in Light, Clean \& Place, Heat \& Place, Cool \& Place, Pick Two \& Place. The calibration subset of 100 episodes was randomly sampled from \textit{valid\_seen} split.

    \subsection{Minigrid}
    \label{appendix:minigrid_additional_details}
    
    \begin{table*}[b]
\centering
\small
\begin{tabular}{l|c}
    \toprule
    \textbf{Model} & \textbf{Success Rate} \\
    \midrule
    \qwensmall    & 0.630 \\
    \llamafour    & 0.105 \\
    \llamathree   & 0.000 \\
    GPT-OSS-20B   & 0.000 \\
    Mistral-8x-7B & 0.000 \\
    Mistral-Small & 0.000 \\
    Qwen3.5-9B    & 0.000 \\
    \midrule
    \gpt          & 0.865 \\
    \qwenlarge    & 0.215 \\
    \qwenmedium   & 0.201 \\
    \bottomrule
\end{tabular}
\caption{Success rate on MiniGrid partial-view across small and large models.}
\label{tab:minigrid_base_partial_view}
\end{table*}
    For MiniGrid, we adopt full-view mode of \textit{MiniGrid-DoorKey-8x8-v0} environment with random seed 42. This environment has a key that the agent must pick up to unlock a door and then get to the goal square to successfully finish the episode. Calibration is performed under the same settings, but with a different random seed (993) to ensure no overlap with the test set. For threshold calibration, we run 100 episodes using this seed. 
    
    We conducted a preliminary investigation in the partial-view settings. However, both the large and small models performed poorly in this regime, so we decided to focus on the full-view mode in the main results. The results for the partial-view mode are presented in Table~\ref{tab:minigrid_base_partial_view}.

    \subsection{Models}
    \label{appendix:models_additional_details}
    
    We report the full model names for all models inferred via Together AI: small models - \textit{Qwen/Qwen3-Next-80B-A3B-Instruct}, \textit{meta-llama/Llama-3.3-70B-Instruct-Turbo} and \textit{meta-llama/Llama-4-Maverick-17B-128E-Instruct-FP8}, large models - \textit{Qwen/Qwen3-235B-A22B-Instruct-2507-tput} and \textit{Qwen/Qwen3-Coder-480B-A35B-Instruct-FP8}.

\section{Metrics}
\label{appendix:metrics}

\paragraph{Prediction Rejection Ratio.}
To evaluate the discriminative quality of UQ methods, we use the Prediction Rejection Ratio (PRR), which measures the effectiveness of uncertainty scores for identifying high-quality predictions~\citep{malinin2021uncertaintyestimationautoregressivestructured}.
PRR operates by progressively rejecting predictions with uncertainty scores above a threshold and observing how the average quality of the remaining predictions changes.
To compute PRR, one needs to estimate areas under two different curves. The first is the area between the Prediction Rejection (PR) curve for the evaluated uncertainty score and a random baseline, and the second is the area between the oracle (the ideal uncertainty score that perfectly aligns with quality) and the random baseline.
Formally, PRR is defined as follows:
\begin{equation}
  \mathrm{PRR} = \frac{\mathrm{AUC}_{\mathrm{unc}} - \mathrm{AUC}_{\mathrm{rnd}}}{\mathrm{AUC}_{\mathrm{oracle}} - \mathrm{AUC}_{\mathrm{rnd}}}.
\end{equation}

\begin{figure}[t!]
    \centering
    \includegraphics[trim={0.cm 0.cm 0.cm 0.cm},clip,width=0.6\linewidth]{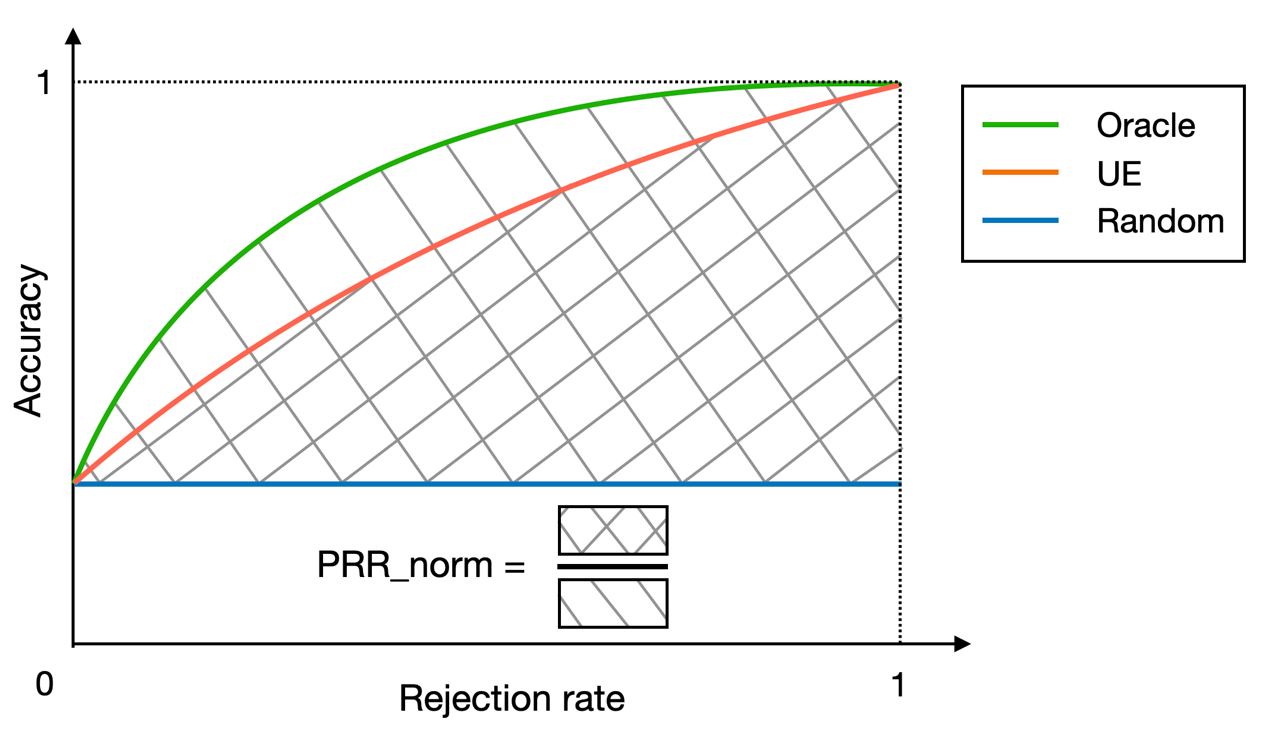}
    \caption{\textbf{Prediction-Rejection Ratio (PRR) Curve.} The curve illustrates the quality of the non-rejected predictions as a function of the rejection rate. \textit{Oracle} represents the optimal rejection strategy, \textit{Random} is a random rejection, and \textit{UE} is rejection based on the evaluated uncertainty estimation method.}
    \label{fig:prr}
  \end{figure}

The higher PRR values indicate better ordering, induced by the uncertainty scores, and approach the performance of an oracle.
To prevent cases where excessive rejection artificially inflates quality measures, we compute PRR only up to a rejection threshold of 50\%.
Figure~\ref{fig:prr} gives a visual representation of the PRR calculation, highlighting the relationship between the uncertainty threshold and the quality measure.

\section{Action vs Reasoning-level Uncertainty Results}
\label{appendix:full_uq_results}

We provide full results for the action-level UQ in Table~\ref{tab:uq_alfworld_action_by_step_full}. The considered UQ methods can be grouped into two main categories.
\begin{itemize}
    \item Information-based: Sequence Probability, Mean Token Entropy and Perplexity.
    
    \item Sample diversity: Number of Semantic Sets~\citep{lingenerating}, Sum of Eigenvalues~\citep{lingenerating}, Eccentricity~\citep{lingenerating}, Lexical Similarity~\citep{fomicheva-etal-2020-unsupervised}, Sentence SAR~\citep{duan-etal-2024-shifting}, Degree Matrix~\citep{lingenerating}, Semantic Density~\citep{qiu2024semanticdensityuncertaintyquantification}, Predictive Entropy~\citep{kuhn2023semantic} and CoCoA~\citep{vashurin2025uncertainty}.
\end{itemize}

For implementation, we use the LM-Polygraph library~\citep{fadeeva2023lmpolygraphuncertaintyestimationlanguage}. To collect data, we sample 100 tasks from ALFWorld and simulate 100 trajectories with the ReAct framework using \qwensmall as the small model, resulting in a total of 2411 steps. Each step was labeled by \gpt as an LLM-as-a-judge, with 1555 steps marked correct and 856 incorrect.

\begin{table*}[t]
\centering
    \begin{tabular}{l|cc}
        \toprule
        \textbf{UQ metric} & \textbf{PRR} & \textbf{ROC-AUC} \\
        \midrule
        Number Semantic Sets & 0.144 & 0.545 \\
        Sum of Eigenvalue & 0.213 & 0.575 \\
        Eccentricity & 0.218 & 0.578 \\
        Lexical Similarity & 0.219 & 0.579 \\
        Sentence SAR & 0.260 & 0.603 \\
        Cocoa PPL & 0.300 & 0.628 \\
        Degree Matrix & 0.334 & 0.668 \\
        Semantic Density & 0.343 & 0.671 \\
        Predictive Entropy & 0.356 & 0.663 \\
        Cocoa SP & 0.357 & 0.662 \\
        Cocoa MTE & 0.363 & 0.664 \\
        Semantic Entropy & 0.365 & 0.665 \\
        \midrule
        Perplexity & 0.399 & 0.684 \\
        Sequence Probability & 0.417 & 0.704 \\
        Mean Token Entropy & \textbf{0.424} & \textbf{0.710} \\
        \bottomrule
    \end{tabular}
\caption{Performance comparison of uncertainty metrics estimated over action selection.}
\label{tab:uq_alfworld_action_by_step_full}
\end{table*}

\section{Prompts}
\label{appendix:prompts}

For reproducibility purposes, we report all prompts used in our experiments:
\begin{itemize}
    \item \textbf{ALFWorld}. The reasoning prompt is displayed in Figure~\ref{fig:alfworld_reasoning_prompt} and the action selection prompt in Figure~\ref{fig:alfworld_action_prompt}. 
    \item \textbf{MiniGrid}. The reasoning prompt is shown in Figure~\ref{fig:minigrid_reasoning_prompt}, while the action selection prompt is provided in Figure~\ref{fig:minigrid_action_prompt}.
\end{itemize}

\begin{figure}
    \centering
    \fbox{
        \parbox{0.8\textwidth}{\scriptsize
You are an AI agent solving a task in an interactive environment.\\
TASK DESCRIPTION:\\
\{DESCRIPTION\}\\
\\
ENVIRONMENT HISTORY:\\
\{HISTORY\}\\
\\
AVAILABLE COMMANDS:\\
\{AVAILABLE\_COMMANDS\}\\
\\
Think step by step about the current situation and consider what action to take next.\\
Your thought process:
        }
    }
    \caption{Prompt to elicit reasoning trace in the ReDAct framework for ALFWorld.}
    \label{fig:alfworld_reasoning_prompt}
\end{figure}

\begin{figure}
    \centering
    \fbox{
        \parbox{0.8\textwidth}{\scriptsize
You are an AI agent solving a task in an interactive environment.\\
\\
TASK DESCRIPTION:\\
\{DESCRIPTION\}\\
\\
ENVIRONMENT HISTORY:\\
\{HISTORY\}\\
\\
YOUR CURRENT REASONING:\\
\{THOUGHTS\}\\
\\
AVAILABLE COMMANDS:\\
\{AVAILABLE\_COMMANDS\}\\
\\
OUTPUT RULES:\\
- Output exactly ONE line.\\
- That line must be EXACTLY one of the AVAILABLE COMMANDS.\\
- Do NOT output reasoning, explanation, punctuation, or extra words.\\
\\
Now output your chosen action (one line only):
        }
    }
    \caption{Prompt to select an action in the ReDAct framework for ALFWorld.}
    \label{fig:alfworld_action_prompt}
\end{figure}

\begin{figure}
    \centering
    \fbox{
        \parbox{0.8\textwidth}{\scriptsize
You are an AI agent solving a task in an interactive environment.\\
\\
TASK DESCRIPTION:\\
\{DESCRIPTION\}\\
\\
ENVIRONMENT HISTORY:\\
\{HISTORY\}\\
\\
AVAILABLE COMMANDS:\\
\{AVAILABLE\_COMMANDS\}\\
\\
Think step by step, but keep your reasoning short, precise, and grounded only in the current observation and history.\\
\\
Follow this reasoning procedure strictly:\\
1. Summarize only the important facts from the current observation and history.\\
2. Identify the current subgoal:\\
   - If the key has not been picked up, the subgoal is to get key.\\
   - If the key has been picked up but the door is not open, the subgoal is to open door.\\
   - If the door is open, the subgoal is to go to goal.\\
3. Identify the exact position of the agent and the current target using the provided grid.\\
   - State the agent position and facing direction.\\
   - State the target position.\\
   - Compute the target's relative location from the agent: same cell, directly ahead, behind, left, right, or requiring movement.\\
   - Determine whether the target is immediately interactable:\\
     - pickup is valid only if the key is in the cell directly in front.\\
     - toggle is valid only if the door is in the cell directly in front.\\
4. Check whether the last few actions show useless repetition or left-right oscillation.\\
5. Choose the one action that makes the most direct progress toward the current subgoal.\\
6. Do not change subgoal unless the current one is completed or impossible.\\
\\
Reasoning rules:\\
- Turning changes orientation only; it does not move the agent.\\
- pickup and toggle work only when the target object is directly in front of the agent.\\
- Do not alternate left and right repeatedly unless there is a clear new reason.\\
- If you already know the door exists but do not have the key, prioritize getting the key instead of returning attention to the door.\\
\\
After forming your initial reasoning, perform a brief self-critique:\\
\\
Critique checklist:\\
- Did I use correct coordinates from the grid?\\
- If I chose pickup or toggle, is the target actually in the front cell?\\
- Does this action make progress toward the current subgoal?\\
- Am I repeating recent actions (e.g., left-right oscillation)?\\
- Am I mistakenly focusing on the door before getting the key?\\
\\
If any issue is found, correct your reasoning and update the action choice.\\
\\
Keep the critique short and focused. Do not repeat the full reasoning.\\
\\
Keep the reasoning compact and action-oriented.\\
Your thought process:
        }
    }
    \caption{Prompt to elicit reasoning trace in the ReDAct framework for MiniGrid.}
    \label{fig:minigrid_reasoning_prompt}
\end{figure}

\begin{figure}
    \centering
    \fbox{
        \parbox{0.8\textwidth}{\scriptsize
You are an AI agent solving a task in an interactive environment.\\
\\
TASK DESCRIPTION:\\
\{DESCRIPTION\}\\
\\
ENVIRONMENT HISTORY:\\
\{HISTORY\}\\
\\
YOUR CURRENT REASONING:\\
\{THOUGHTS\}\\
\\
AVAILABLE COMMANDS:\\
\{AVAILABLE\_COMMANDS\}\\
\\
OUTPUT RULES:\\
- Output exactly ONE line.\\
- That line must be EXACTLY one of the AVAILABLE COMMANDS.\\
- Do NOT output reasoning, explanation, punctuation, or extra words.\\
\\
Now output your chosen action (one line only):
        }
    }
    \caption{Prompt to select an action in the ReDAct framework for MiniGrid.}
    \label{fig:minigrid_action_prompt}
\end{figure}

\begin{figure}
    \centering
    \fbox{
        \parbox{0.8\textwidth}{\scriptsize
You are an expert evaluator of agent trajectories in ALFWorld.\\
\\
Your task:\\
Evaluate whether each step in the trajectory is helpful toward completing the task.\\
\\
Output format:\\
Return ONLY a single JSON object.\\
Each key must be "step i" (with i starting from 1).\\
Each value must be an object with:\\
- "label": 0 or 1\\
- "reason": a short 1 sentence explanation\\
\\
Example structure:\\
{\\
  "step 1": { "label": 1, "reason": "Going to the fridge is a reasonable way to look for food items such as eggs." },\\
  "step 2": { "label": 0, "reason": "This repeats a previous inspection without adding new information." }\\
}\\
\\
Do NOT include any extra text or code fences outside the JSON.\\
Return ONLY the raw JSON.\\
\\
Labeling rules:\\
\\
A step should be labeled 1 (good) if:\\
- It is a reasonable attempt to find the needed object for the task.\\
- It involves searching a plausible location (e.g., opening a cabinet, drawer, fridge, etc.).\\
- It examines or moves toward objects likely relevant to the task or the next logical subgoal.\\
- It is part of logical exploration, even if the final goal is not yet achieved.\\
\\
A step should be labeled 0 (bad) if:\\
- It repeats an action with no new benefit (e.g., re-examining the same object without a clear reason).\\
- It returns to a place already fully explored without purpose.\\
- It manipulates objects clearly irrelevant to the task.\\
- It represents looping or wandering behavior that does not help progress.\\
\\
Important:\\
- Exploration steps SHOULD be labeled 1 if they could reasonably lead to the target object.\\
- Do NOT mark all steps as 0 unless literally every step is clearly irrelevant to the task.\\
\\
Now evaluate the following trajectory.\\
Return ONLY the JSON.\\
\\
Trajectory:\\
\{TRAJECTORY\}\\
        }
    }
    \caption{Prompt for \gpt to label steps in AlfWorld trajectories.}
    \label{fig:gpt_judge_prompt}
\end{figure}

\section{Additional Experimental Results}
\label{appendix:additional_experimental_results}

\paragraph{Number of calls to large model.}
Tables~\ref{tab:alfworld_mean_oracle_calls_per_episode} and~\ref{tab:minigrid_mean_oracle_calls_per_episode} report the number of calls to a large model for the ALFWorld and MiniGrid benchmarks, respectively. Approximate parity has been achieved between various setups, with the exception of \llamathree runs on ALFWorld, where the random baseline used slightly fewer large model calls. On the other hand, in MiniGrid experiments with \qwensmall as the small model, the random baseline used slightly more calls. We attribute this to the limitations of the calibration procedure, discussed in detail in Section~\ref{sec:threshold}.

\begin{figure}[t!]
    \centering
    \begin{subfigure}[c]{\linewidth}
    \includegraphics[trim={0.cm 0.cm 0.cm 0.cm},clip,width=0.99\linewidth]{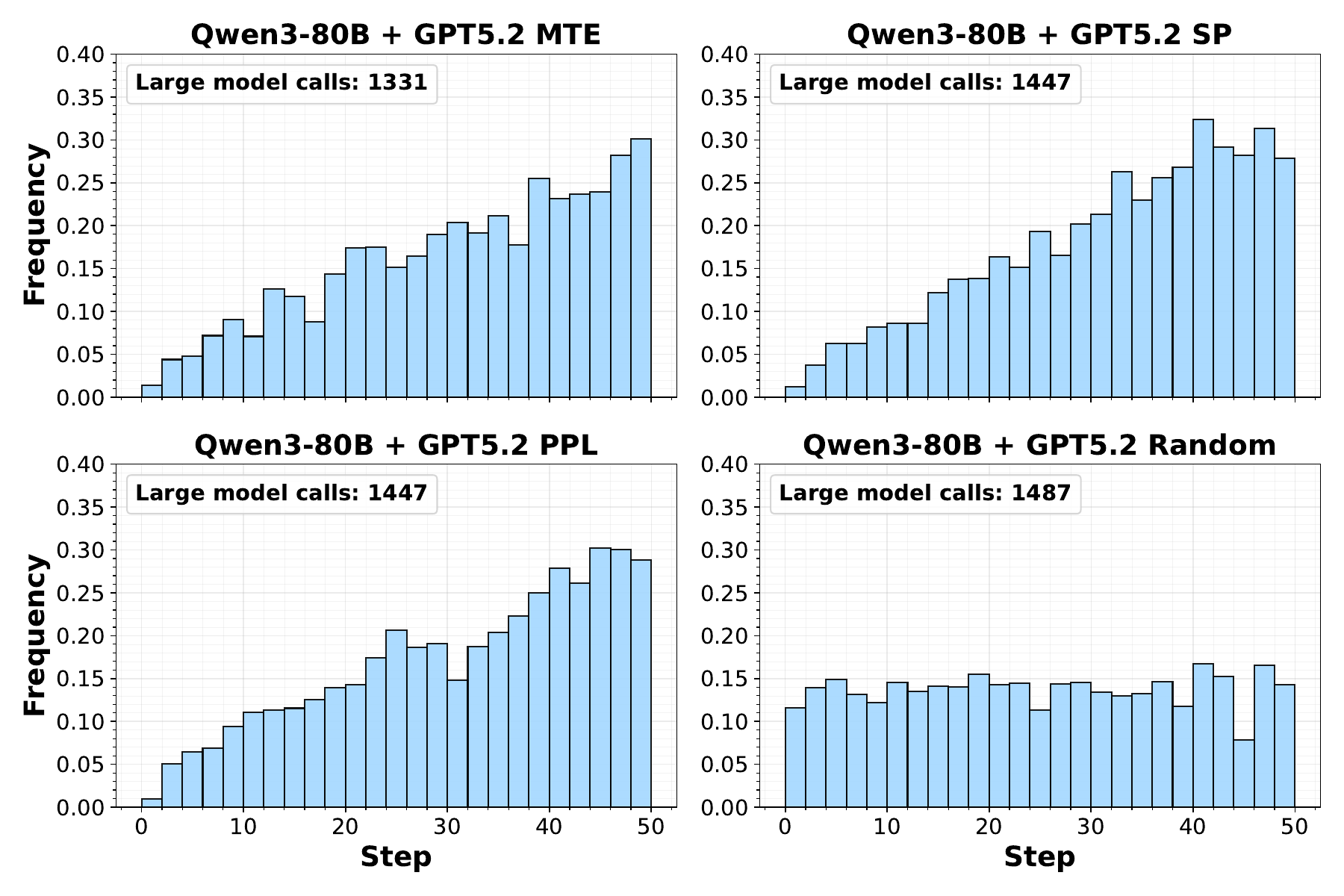}
    \caption{\textbf{\qwensmall + \gpt.} Large model invocation frequency by step in ALFWorld. At each step, the frequency is defined as the number of large model calls divided by the number of episodes that reached that step.}
    \label{fig:num_oracle_calls_alfworld_qwen_gpt}
    \end{subfigure}

    \begin{subfigure}[c]{\linewidth}
    \centering
    \includegraphics[trim={0.cm 0.cm 0.cm 0.cm},clip,width=0.99\linewidth]{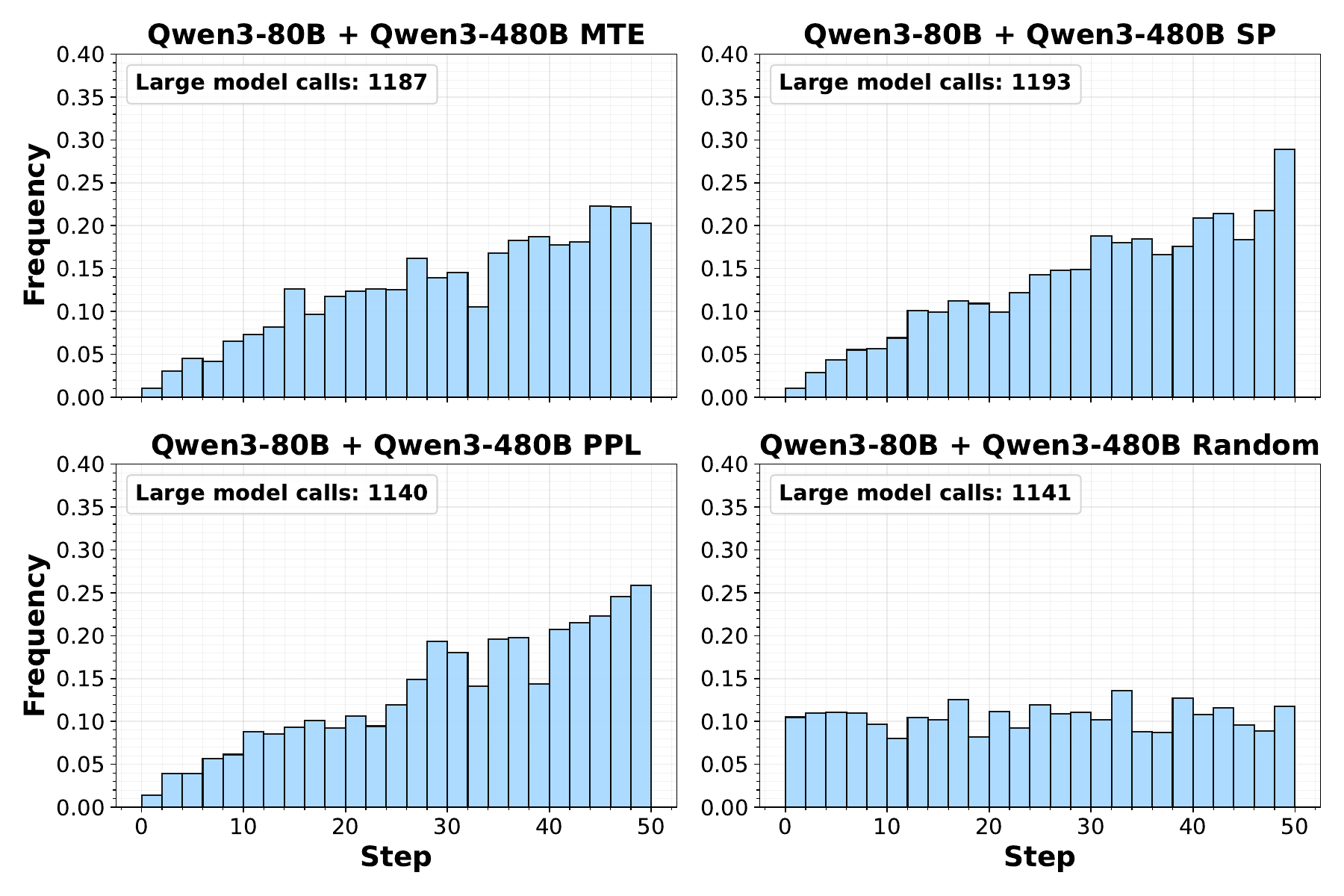}
    \caption{\textbf{\qwensmall + \qwenlarge.} Large model invocation frequency by step in ALFWorld. At each step, the frequency is defined as the number of large model calls divided by the number of episodes that reached that step.}
    \label{fig:num_oracle_calls_alfworld_qwen_qwen}
  \end{subfigure}
  \end{figure}

\begin{figure}[t!]
    \begin{subfigure}[c]{\linewidth}
    \centering
    \includegraphics[trim={0.cm 0.cm 0.cm 0.cm},clip,width=0.99\linewidth]{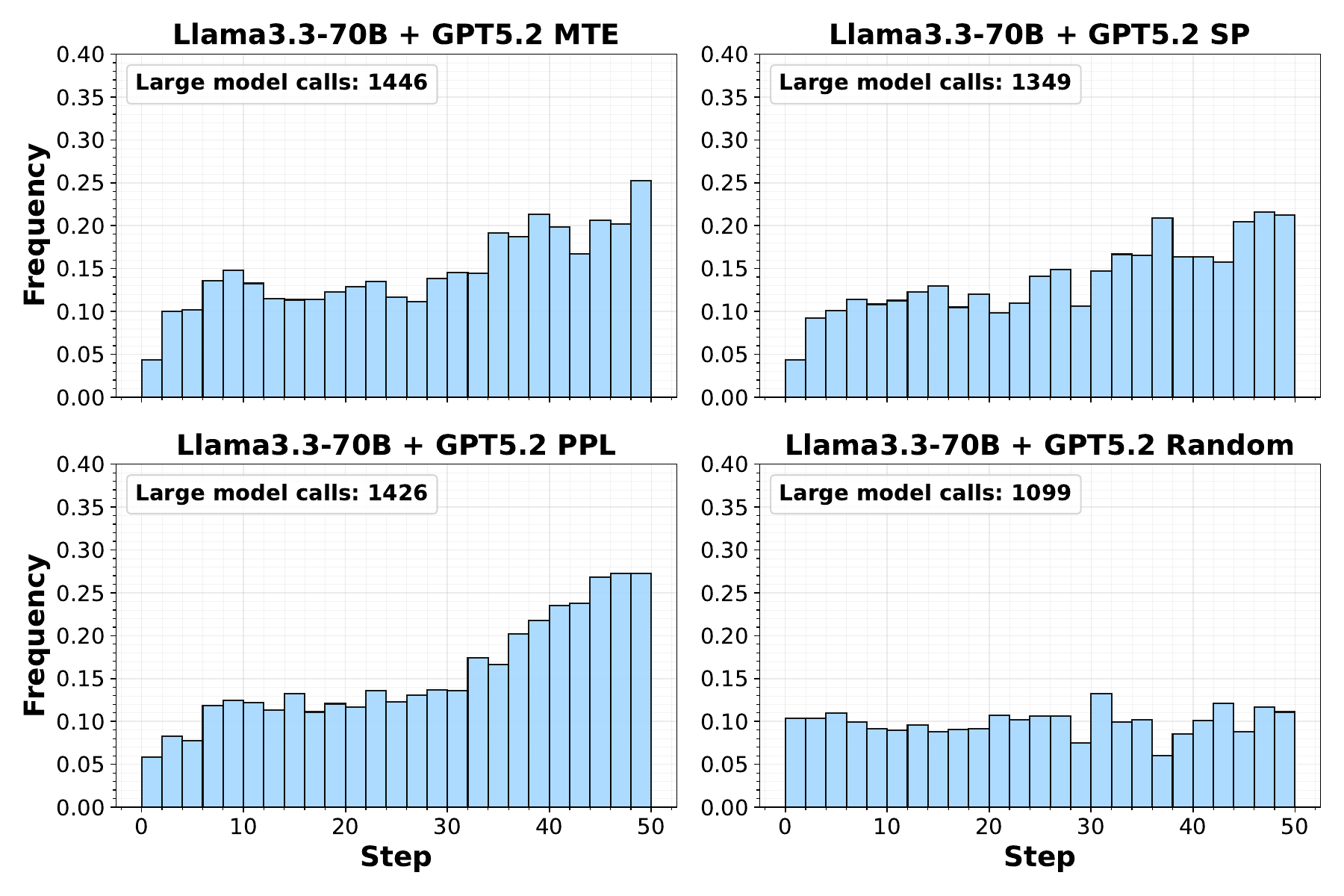}
    \caption{\textbf{\llamathree + \gpt.} Large model invocation frequency by step in ALFWorld. At each step, the frequency is defined as the number of large model calls divided by the number of episodes that reached that step.}
    \label{fig:num_oracle_calls_alfworld_llama_gpt}
  \end{subfigure}

  \begin{subfigure}[c]{\linewidth}
    \centering
    \includegraphics[trim={0.cm 0.cm 0.cm 0.cm},clip,width=0.99\linewidth]{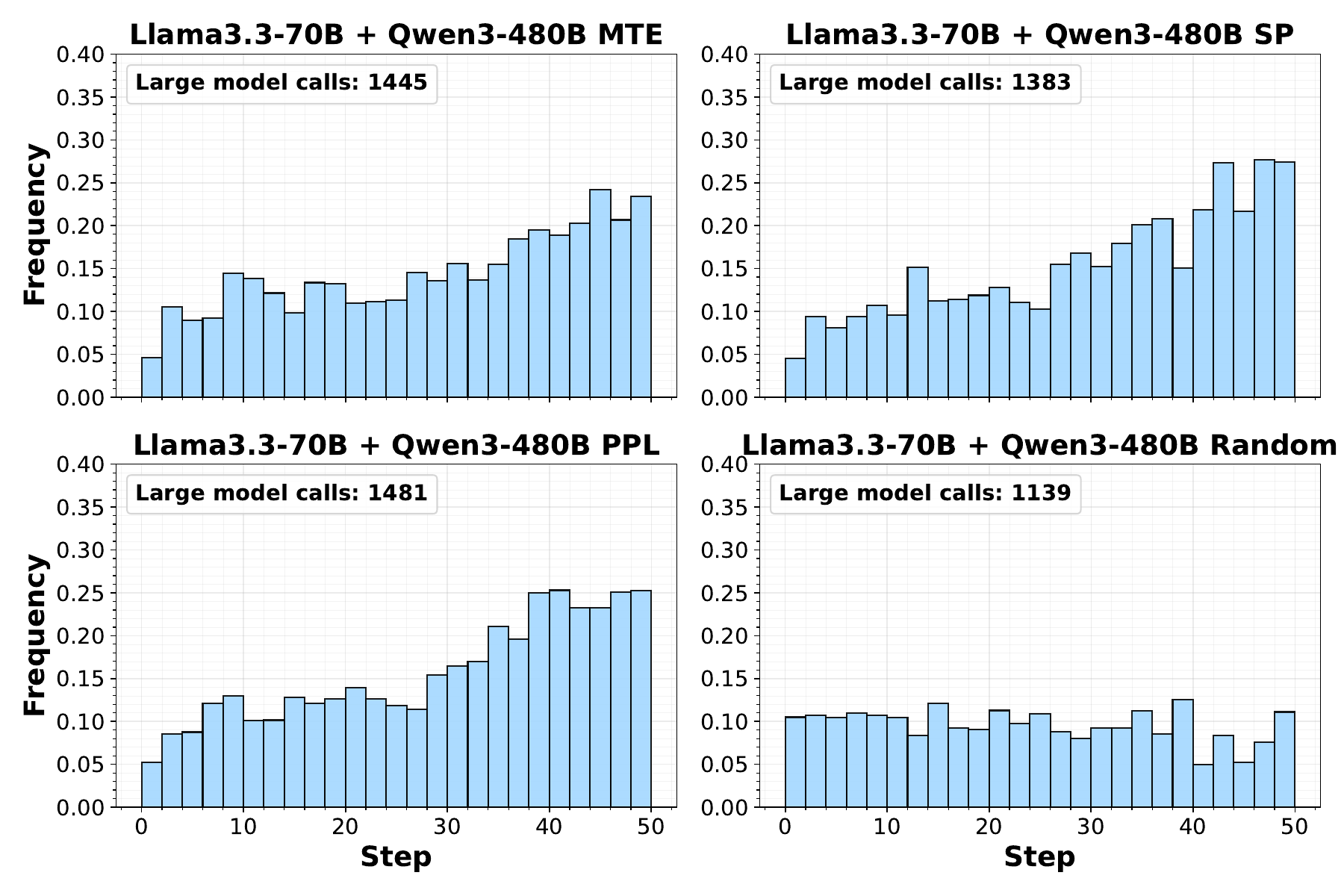}
    \caption{\textbf{\llamathree + \qwenlarge.} Large model invocation frequency by step in ALFWorld. At each step, the frequency is defined as the number of large model calls divided by the number of episodes that reached that step.}
    \label{fig:num_oracle_calls_alfworld_llama_qwen}
  \end{subfigure}
  \end{figure}

\begin{figure}[t!]
    \begin{subfigure}[c]{\linewidth}
    \centering
    \includegraphics[trim={0.cm 0.cm 0.cm 0.cm},clip,width=0.99\linewidth]{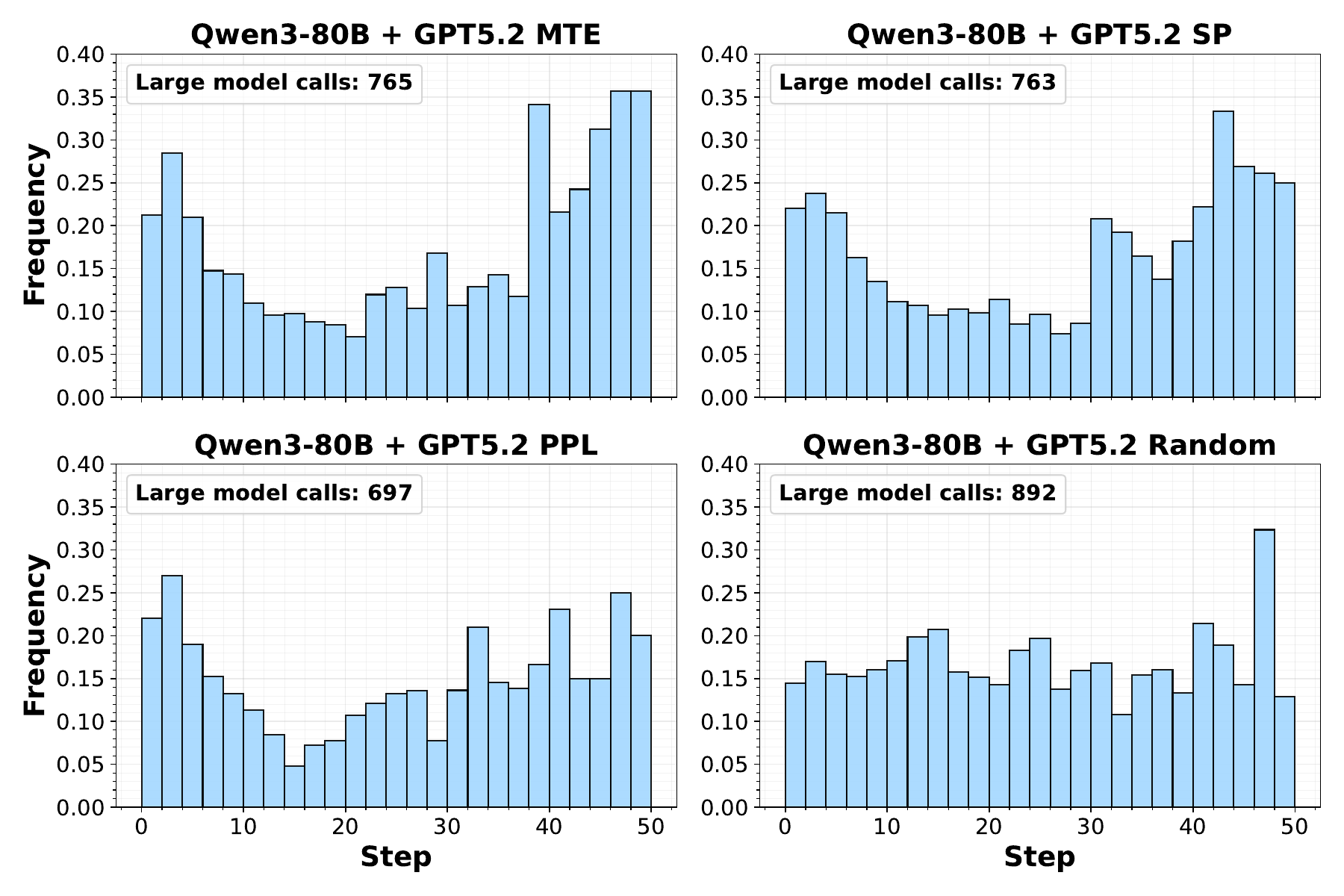}
    \caption{\textbf{\qwensmall + \gpt.} Large model invocation frequency by step in MiniGrid. At each step, the frequency is defined as the number of large model calls divided by the number of episodes that reached that step.}
    \label{fig:num_oracle_calls_minigrid_qwen_gpt}
  \end{subfigure}

  \begin{subfigure}[c]{\linewidth}
    \centering
    \includegraphics[trim={0.cm 0.cm 0.cm 0.cm},clip,width=0.99\linewidth]{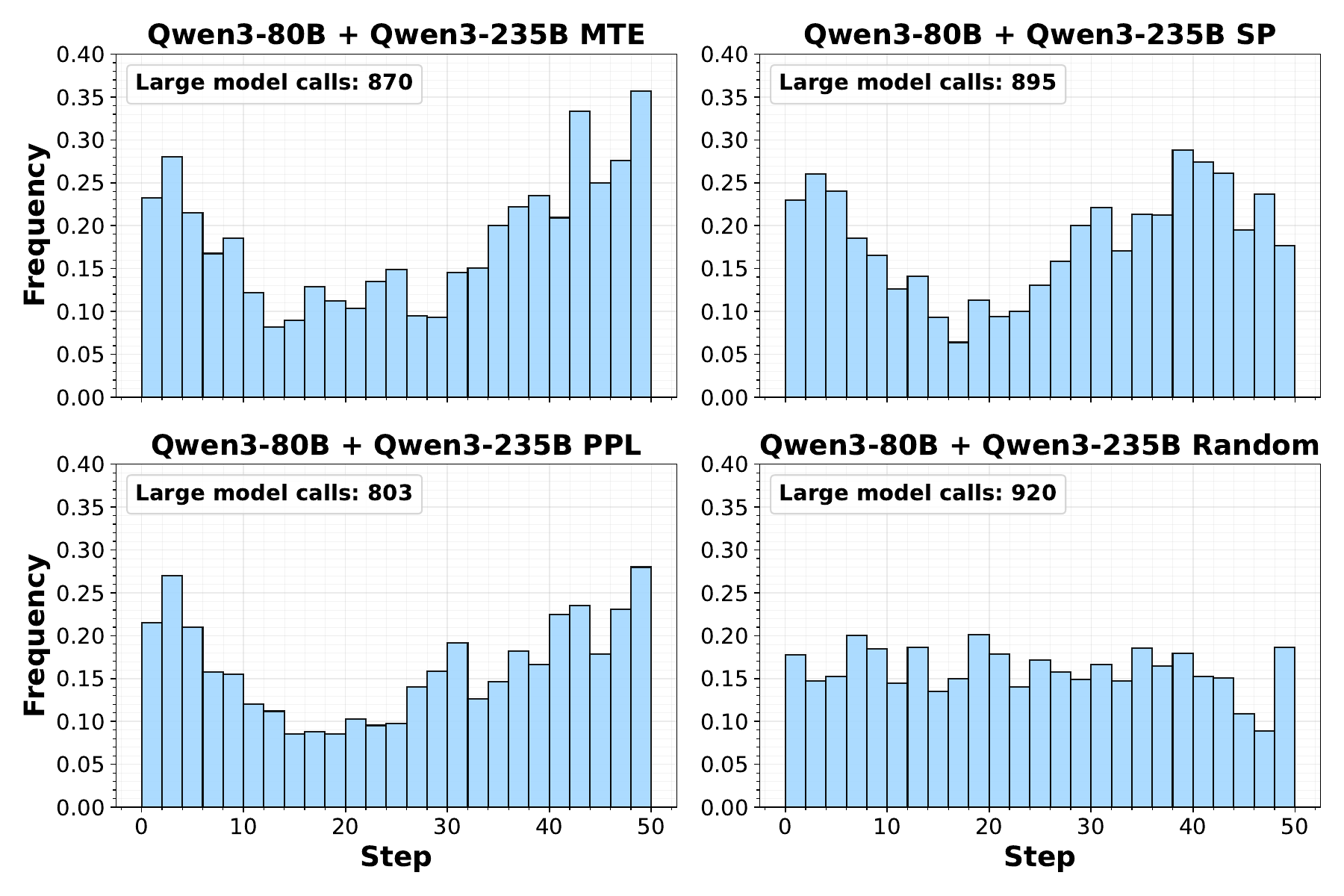}
    \caption{\textbf{\qwensmall + \qwenlarge.} Large model invocation frequency by step in MiniGrid. At each step, the frequency is defined as the number of large model calls divided by the number of episodes that reached that step.}
    \label{fig:num_oracle_calls_minigrid_qwen_qwen}
    \end{subfigure}
  \end{figure}

\begin{figure}[t!]
  \begin{subfigure}[c]{\linewidth}
    \centering
    \includegraphics[trim={0.cm 0.cm 0.cm 0.cm},clip,width=0.99\linewidth]{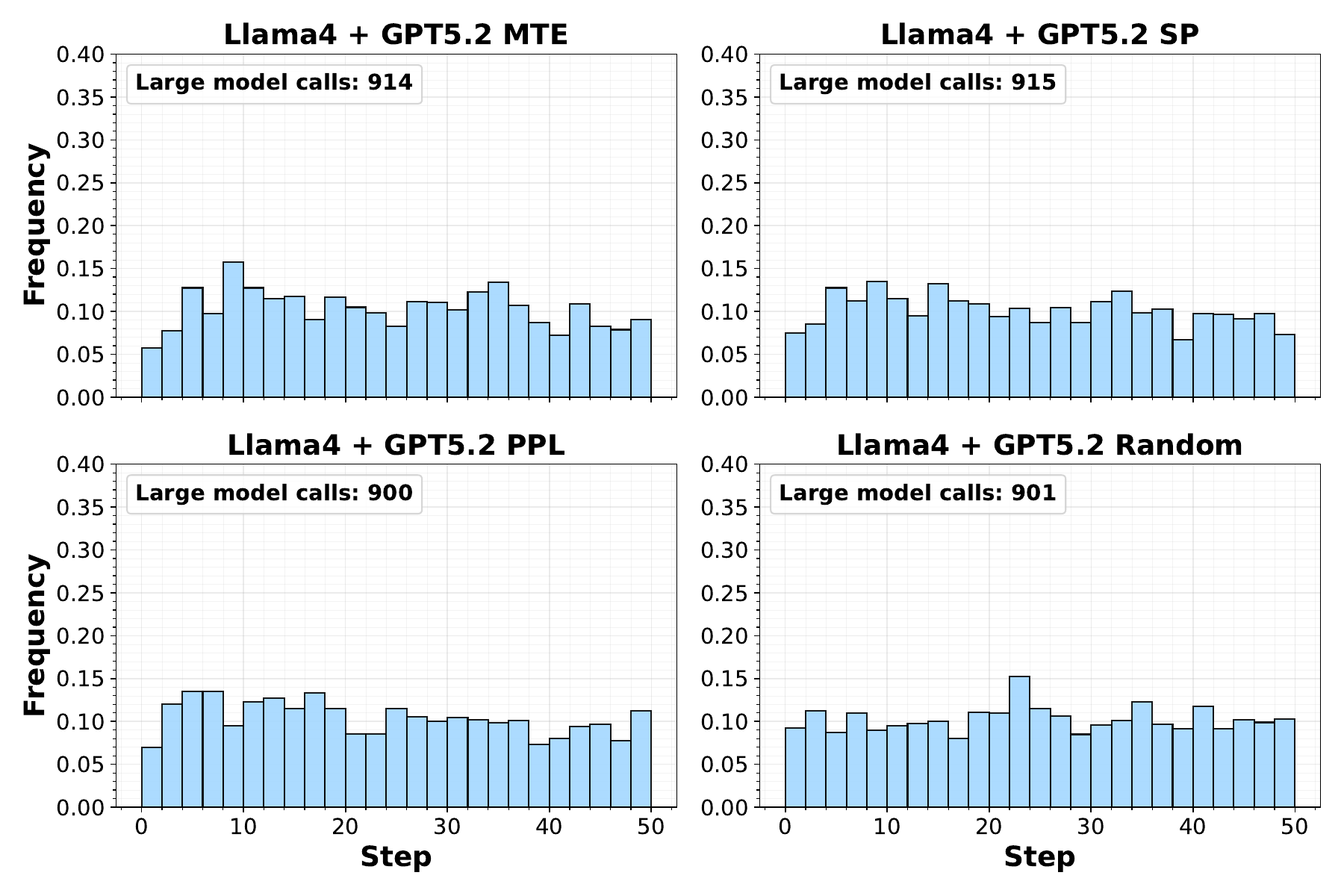}
    \caption{\textbf{\llamafour + \gpt.} Large model invocation frequency by step in MiniGrid. At each step, the frequency is defined as the number of large model calls divided by the number of episodes that reached that step.}
    \label{fig:num_oracle_calls_minigrid_llama_gpt}
  \end{subfigure}

  \begin{subfigure}[c]{\linewidth}
    \centering
    \includegraphics[trim={0.cm 0.cm 0.cm 0.cm},clip,width=0.99\linewidth]{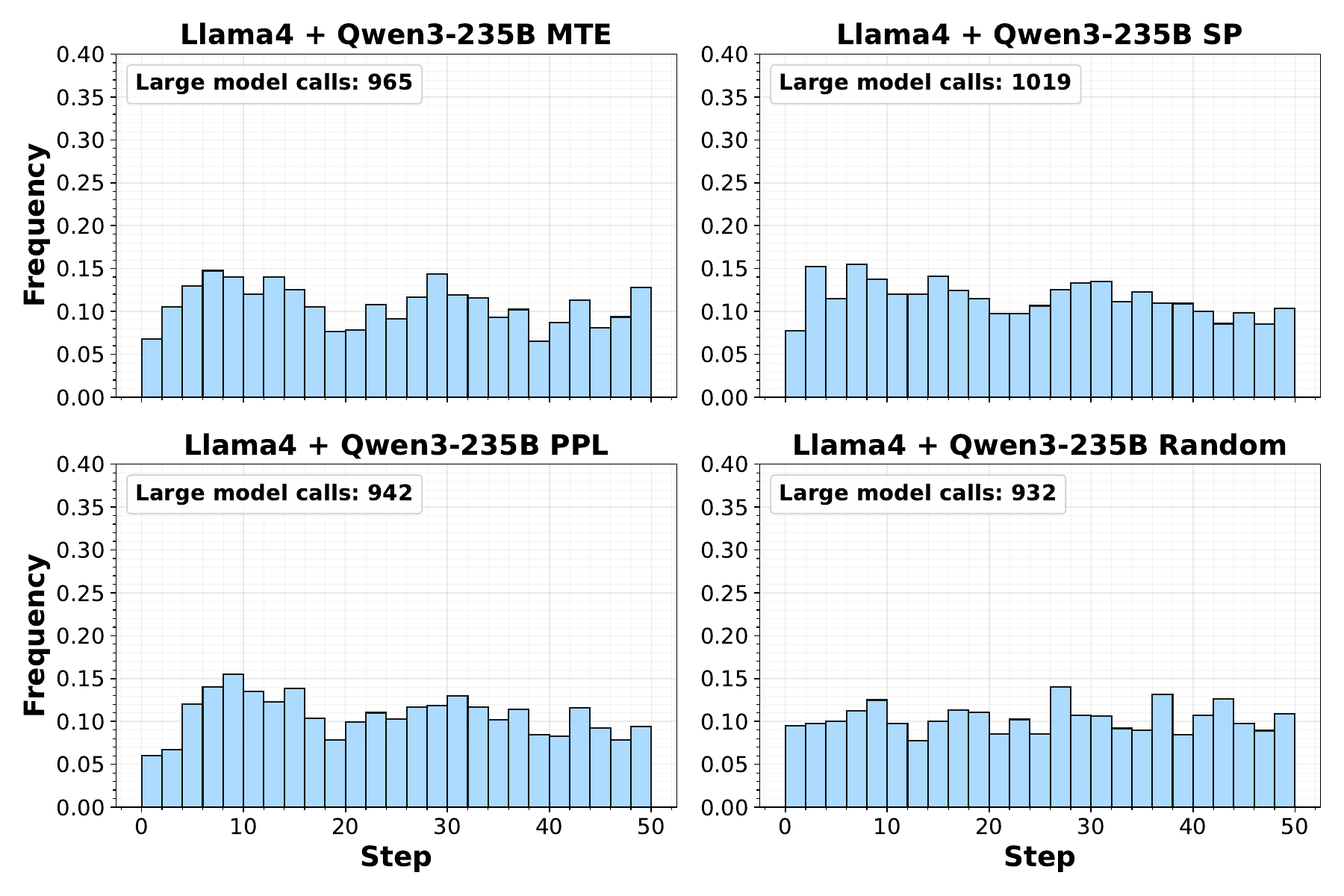}
    \caption{\textbf{\llamafour + \qwenlarge.} Large model invocation frequency by step in MiniGrid. At each step, the frequency is defined as the number of large model calls divided by the number of episodes that reached that step.}
    \label{fig:num_oracle_calls_minigrid_llama_qwen}
  \end{subfigure}
  \end{figure}

\paragraph{Distributions of calls to large model.}
Figures~\ref{fig:num_oracle_calls_alfworld_qwen_gpt} through~\ref{fig:num_oracle_calls_minigrid_llama_qwen} depict the distributions of the large model calls for the remainder of the experimental setups. The general characteristics follow the patterns discussed in Section~\ref{sec:experiments}, with a notable exception of experiments where \llamafour acts as the small model on the MiniGrid benchmark. Here, UQ-guided deferral is distributed closer to a uniform distribution than in other setups. The low base quality of this model on MiniGrid can explain this - the model is bad enough for the task to not know when to express confidence, which leads to a flatter distribution of calls.

\paragraph{Number of steps to success.}
Tables~\ref{tab:alfworld_number_steps_to_success} and~\ref{tab:minigrid_number_steps_to_success} report average number of steps until completion for successful episodes for ALFWorld and MiniGrid, respectively. Given the magnitude of standard deviations, we conclude that deferral does not significantly impact the expected completion time for both benchmarks.

\paragraph{Token usage.}
Tables~\ref{tab:alfworld_detailed_tokens_qwen_gpt} through~\ref{tab:alfworld_detailed_tokens_llama_qwen} report detailed breakdown of token usage along with associated costs for all experiments on ALFWorld.

\begin{figure}[t!]
    \centering
    \begin{minipage}{0.49\linewidth}
        \centering
        \includegraphics[width=\linewidth]{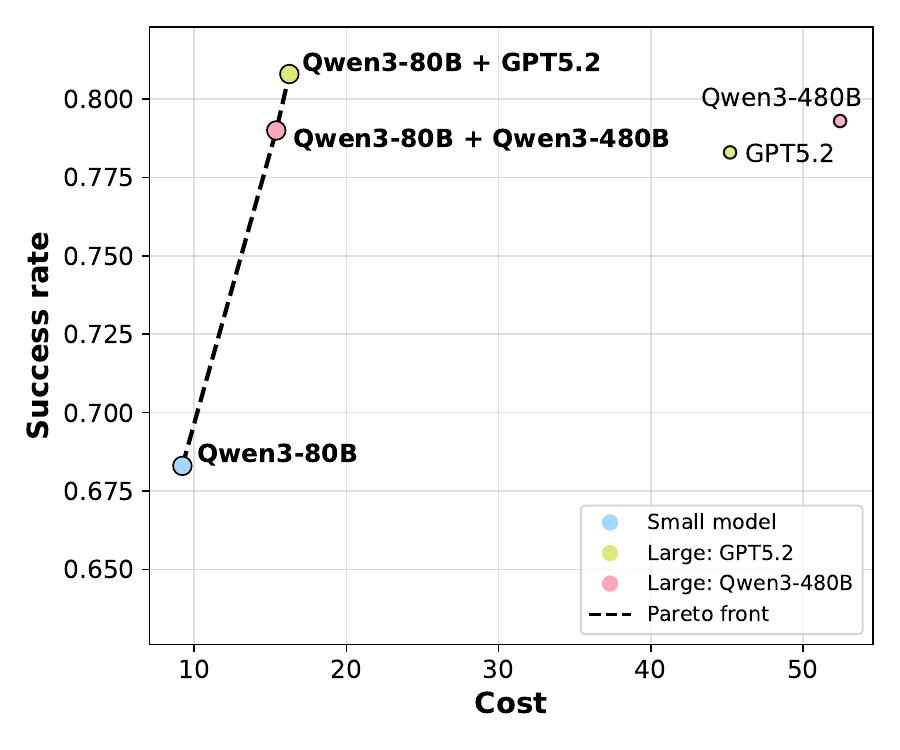}
        {\small (a) \qwensmall}
    \end{minipage}
    \hfill
    \begin{minipage}{0.49\linewidth}
        \centering
        \includegraphics[width=\linewidth]{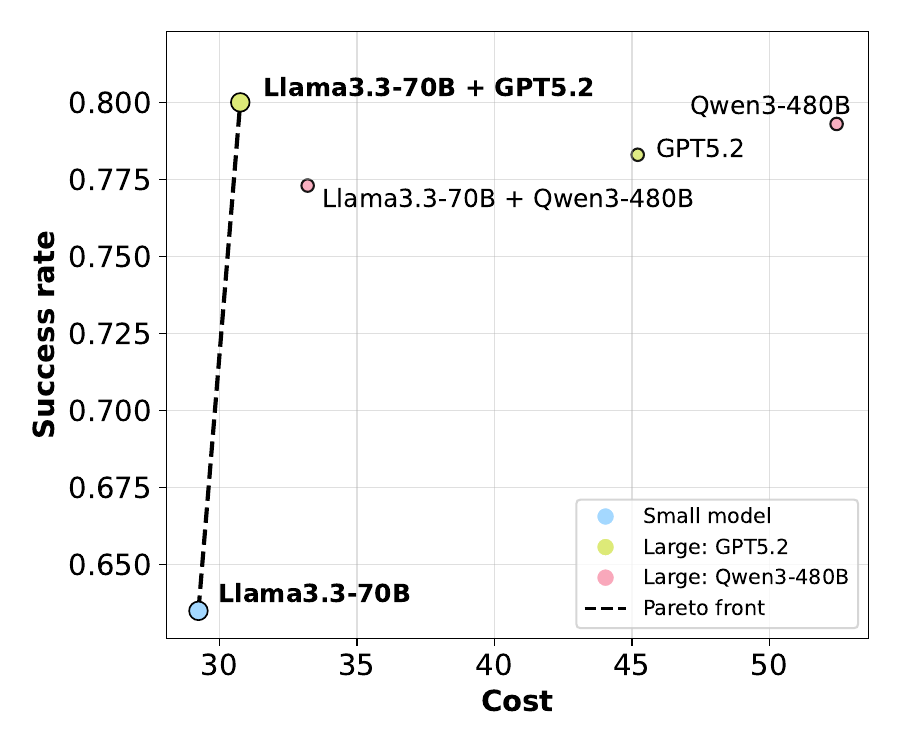}
        {\small (b) \llamathree}
    \end{minipage}

    \caption{\textbf{Pareto front of success rate vs.\ cost on ALFWorld with perplexity-based deferral.}
    Each plot shows the trade-off between performance and computational cost. 
    (a): \qwensmall\ small model. (b): \llamathree\ small model.}
    \label{fig:pareto_alfworld_cost_combined}
\end{figure}

\begin{figure}[t!]
    \centering
    \begin{minipage}{0.49\linewidth}
        \centering
        \includegraphics[width=\linewidth]{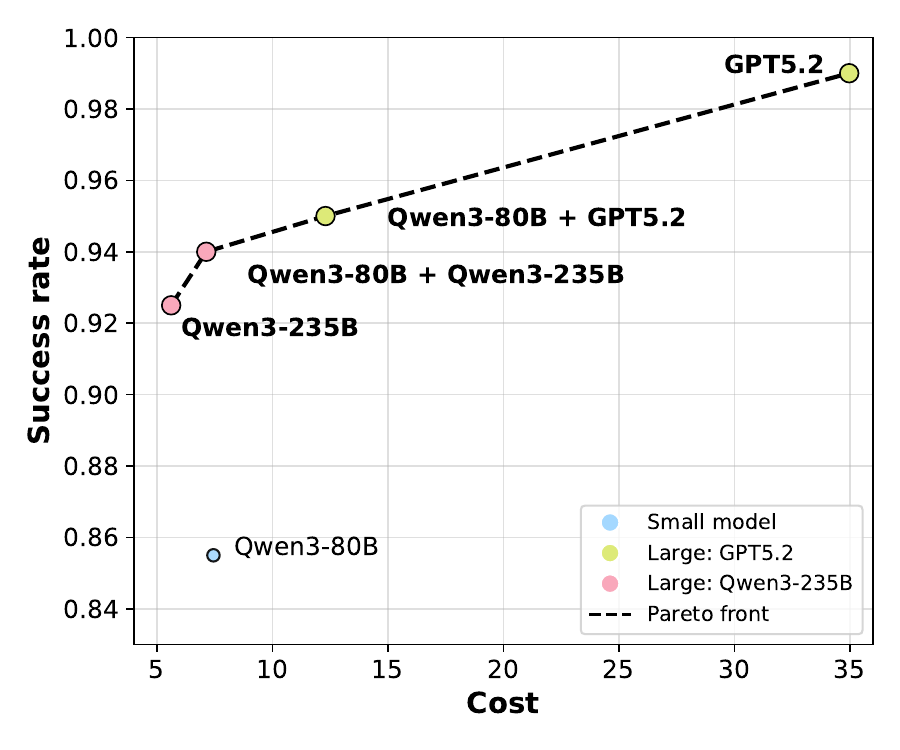}
        {\small (a) \qwensmall}
    \end{minipage}
    \hfill
    \begin{minipage}{0.49\linewidth}
        \centering
        \includegraphics[width=\linewidth]{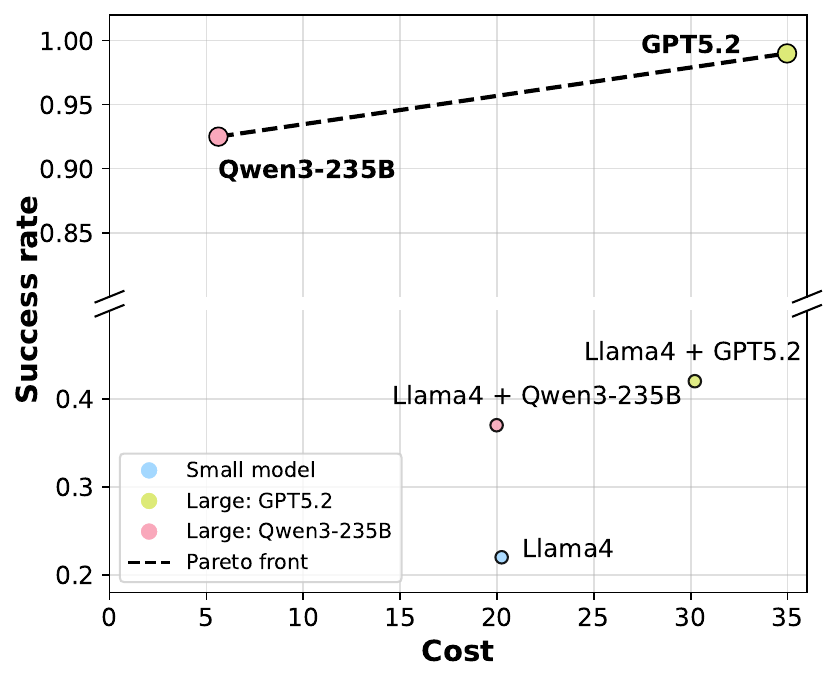}
        {\small (b) \llamafour}
    \end{minipage}

    \caption{\textbf{Pareto front of success rate vs.\ cost on MiniGrid with perplexity-based deferral.}
    Each plot shows the trade-off between performance and computational cost. 
    (a): \qwensmall\ small model. (b): \llamafour\ small model.}
    \label{fig:pareto_minigrid_cost_combined}
\end{figure}

\paragraph{Pareto-optimality with respect to cost.}
Figures~\ref{fig:pareto_alfworld_cost_combined} and~\ref{fig:pareto_minigrid_cost_combined} report Pareto-optimality of ReDAct with respect to success rate and inference cost in USD. Almost all ReDAct variants lie on the Pareto front, with the exception of \llamafour on MiniGrid, where ReDAct lies behind the front due to its low base performance on the benchmark. Still, ReDAct doubles the small model's success rate at only a modest cost increase.

%%%%%%%%%%%%%%%
\begin{table}[!t]
\centering
\begin{tabular}{lcc}
    \toprule
    \textbf{Model} & \textbf{Input Tokens Price} & \textbf{Output Tokens Price} \\
    \midrule
    \qwensmall     & 0.15  & 1.50 \\
    \llamathree    & 0.88  & 0.88 \\
    \llamafour     & 0.27  & 0.85 \\
    \midrule
    \gpt         & 1.75  & 14.00 \\
    \qwenlarge     & 2.00  & 2.00 \\
    \bottomrule
\end{tabular}
\caption{Input and Output token price by model in USD per million tokens.}
\label{tab:price_list}
\end{table}
%%%%%%%%%%%%%%%

  \begin{table*}[t]
\centering
\small
\begin{tabular}{ll|cccc}
    \toprule
    \textbf{Base} & \textbf{Setting} & MTE & SP & PPL & Random \\
    \midrule
    
    \multirow{3}{*}{\qwensmall}
        & Base         & \multicolumn{4}{c}{0} \\
        & + \gpt       & 1331 & 1447 & 1447 & 1487 \\
        & + \qwenlarge & 1187 & 1193 & 1140 & 1141 \\
    
    \midrule
    
    \multirow{3}{*}{\llamathree}
        & Base         & \multicolumn{4}{c}{0} \\
        & + \gpt       & 1446 & 1349 & 1426 & 1099 \\
        & + \qwenlarge & 1445 & 1383 & 1481 & 1139 \\
    
    \midrule
    \gpt       & Base & \multicolumn{4}{c}{9898} \\
    \qwenlarge & Base & \multicolumn{4}{c}{10425} \\
    
    \bottomrule
\end{tabular}
\caption{Number of large model calls on ALFWorld. For the small model without deferral, all methods coincide.}
\label{tab:alfworld_oracle_calls}
\end{table*}
  \begin{table*}[t]
\centering
\small
\begin{tabular}{ll|cccc}
    \toprule
    \textbf{Base} & \textbf{Setting} & MTE & SP & PPL & Random \\
    \midrule
    
    \multirow{3}{*}{\qwensmall}
        & Base         & \multicolumn{4}{c}{0} \\
        & + \gpt       & 765 & 763 & 697 & 892 \\
        & + \qwenmedium & 870 & 895 & 803 & 920 \\
    
    \midrule
    
    \multirow{3}{*}{\llamafour}
        & Base         & \multicolumn{4}{c}{0} \\
        & + \gpt       & 914 & 915 & 900 & 901 \\
        & + \qwenmedium & 965 & 1019 & 942 & 932 \\
    
    \midrule
    \gpt       & Base & \multicolumn{4}{c}{3820} \\
    \qwenmedium & Base & \multicolumn{4}{c}{5047} \\
    
    \bottomrule
\end{tabular}
\caption{Number of large model calls on MiniGrid. For the small model without deferral, all methods coincide.}
\label{tab:minigrid_oracle_calls}
\end{table*}
  \begin{table*}[t]
\centering
\small
\begin{tabular}{ll|cccc}
    \toprule
    \textbf{Base} & \textbf{Setting} & MTE & SP & PPL & Random \\
    \midrule
    
    \multirow{3}{*}{\qwensmall}
        & Base         & \multicolumn{4}{c}{19.58 $\pm$ 11.72} \\
        & + \gpt    & 20.31 $\pm$ 12.04 & 20.87 $\pm$ 12.61 & 21.49 $\pm$ 13.41 & 20.36 $\pm$ 12.10 \\
        & + \qwenlarge & 21.39 $\pm$ 12.61 & 20.88 $\pm$ 13.24 & 21.32 $\pm$ 12.78 & 19.96 $\pm$ 12.42 \\
    
    \midrule
    
    \multirow{3}{*}{\llamathree}
        & Base         & \multicolumn{4}{c}{20.83 $\pm$ 13.27} \\
        & + \gpt    & 20.78 $\pm$ 12.68 & 20.47 $\pm$ 12.09 & 21.32 $\pm$ 12.78 & 20.35 $\pm$ 12.43 \\
        & + \qwenlarge & 21.12 $\pm$ 12.34 & 19.44 $\pm$ 11.91 & 21.09 $\pm$ 12.54 & 19.35 $\pm$ 11.45 \\
    
    \midrule
    \gpt & Base     & \multicolumn{4}{c}{17.73 $\pm$ 12.31} \\
    \qwenlarge & Base  & \multicolumn{4}{c}{19.79 $\pm$ 12.19} \\
    
    \bottomrule
\end{tabular}
\caption{Average number of steps before success with standard deviation for ALFWorld.}
\label{tab:alfworld_number_steps_to_success}
\end{table*}

  \begin{table*}[t]
\centering
\small
\begin{tabular}{ll|cccc}
    \toprule
    \textbf{Base} & \textbf{Setting} & MTE & SP & PPL & Random \\
    \midrule
    
    \multirow{3}{*}{\qwensmall}
        & Base & \multicolumn{4}{c}{26.47 $\pm$ 8.62} \\
        & + \gpt & 24.12 $\pm$ 7.09 & 25.02 $\pm$ 7.72 & 24.08 $\pm$ 6.81 & 25.09 $\pm$ 7.61 \\
        & + \qwenmedium & 25.65 $\pm$ 7.55 & 25.34 $\pm$ 7.49 & 25.64 $\pm$ 8.06 & 25.38 $\pm$ 7.81 \\
    
    \midrule
    
    \multirow{3}{*}{\llamafour}
        & Base & \multicolumn{4}{c}{33.41 $\pm$ 9.79} \\
        & + \gpt & 33.53 $\pm$ 7.03 & 33.83 $\pm$ 8.40 & 32.40 $\pm$ 8.76 & 33.93 $\pm$ 8.89 \\
        & + \qwenmedium & 33.17 $\pm$ 8.99 & 32.44 $\pm$ 8.99 & 32.53 $\pm$ 8.80 & 33.49 $\pm$ 9.18 \\
    
    \midrule
    \gpt & Base & \multicolumn{4}{c}{18.79 $\pm$ 4.12} \\
    \qwenmedium & Base  & \multicolumn{4}{c}{23.23 $\pm$ 6.46} \\
    
    \bottomrule
\end{tabular}
\caption{Average number of steps before success with standard deviation for MiniGrid.}
\label{tab:minigrid_number_steps_to_success}
\end{table*}

  \begin{table*}[t]
\centering
\small
\begin{tabular}{ll|cccc}
    \toprule
    \textbf{Base} & \textbf{Setting} & MTE & SP & PPL & Random \\
    \midrule
    
    \multirow{2}{*}{\qwensmall}
        & + \gpt    & 3.328 $\pm$ 5.270 & 3.618 $\pm$ 5.453 & 3.618 $\pm$ 5.369 & 3.717 $\pm$ 2.852 \\
        & + \qwenlarge & 2.967 $\pm$ 4.747 & 2.982 $\pm$ 4.406 & 2.850 $\pm$ 4.405 & 2.853 $\pm$ 2.328 \\
    
    \midrule
    
    \multirow{2}{*}{\llamathree}
        & + \gpt   & 3.615 $\pm$ 3.990 & 3.373 $\pm$ 3.947 & 2.850 $\pm$ 4.405 & 2.748 $\pm$ 2.195 \\
        & + \qwenlarge & 3.612 $\pm$ 4.343 & 3.458 $\pm$ 4.633 & 3.703 $\pm$ 4.638 & 2.848 $\pm$ 2.346 \\
    
    \bottomrule
\end{tabular}
\caption{Mean number of large model calls with standard deviation per episode for ALFWorld.}
\label{tab:alfworld_mean_oracle_calls_per_episode}
\end{table*}

  \begin{table*}[t]
\centering
\small
\begin{tabular}{ll|cccc}
    \toprule
    \textbf{Base} & \textbf{Setting} & MTE & SP & PPL & Random \\
    \midrule
    
    \multirow{2}{*}{\qwensmall}
        & + \gpt & 3.825 $\pm$ 3.449 & 3.815 $\pm$ 3.211 & 3.485 $\pm$ 2.421 & 4.460 $\pm$ 2.414 \\
        & + \qwenmedium & 4.350 $\pm$ 3.594 & 4.475 $\pm$ 3.703 & 4.015 $\pm$ 3.126 & 4.600 $\pm$ 2.532 \\
    
    \midrule
    
    \multirow{2}{*}{\llamafour}
        & + \gpt & 4.570 $\pm$ 2.673 & 4.575 $\pm$ 2.708 & 4.500 $\pm$ 2.706 & 4.505 $\pm$ 2.121 \\
        & + \qwenmedium & 4.825 $\pm$ 2.786 & 5.095 $\pm$ 2.898 & 4.710 $\pm$ 2.531 & 4.660 $\pm$ 2.125 \\
    
    \bottomrule
\end{tabular}
\caption{Mean number of large model calls with standard deviation per episode for MiniGrid.}
\label{tab:minigrid_mean_oracle_calls_per_episode}
\end{table*}

  \begin{table}[ht]
\centering
\begin{tabular}{l|rrrr}
    \toprule
    & MTE & SP & PPL & Random \\
    \midrule
    Small Model Input Tokens
    & 25,638,669 & 25,850,196 & 26,166,777 & 25,463,015 \\
    
    Small Model Output Tokens
    & 2,530,446 & 2,522,110 & 2,546,153 & 2,579,383 \\
    
    Large Model Input Tokens
    & 3,905,271 & 4,372,542 & 4,236,147 & 3,209,024 \\
    
    Large Model Output Tokens
    & 68,350 & 76,765 & 78,514 & 60,285 \\

    Large Model Calls 
    & 1,331 & 1,447 & 1,447 & 1,487 \\

    \midrule
    
    Small Model price \$ 
    & 7.64 & 7.66 & 7.74 & 7.69 \\
    
    Large Model price \$ 
    & 7.79 & 8.73 & 8.51 & 6.46 \\
    
    Total price \$ 
    & 15.43 & 16.39 & 16.25 & 14.15 \\
    \bottomrule
\end{tabular}
\caption{ALFWorld with \qwensmall + \gpt. Token usage and cost comparison across different uncertainty metrics.}
\label{tab:alfworld_detailed_tokens_qwen_gpt}
\end{table}
  \begin{table}[ht]
\centering
\begin{tabular}{l|rrrr}
    \toprule
    & MTE & SP & PPL & Random \\
    \midrule
    Small Model Input Tokens
    & 27,982,528 & 26,776,457 & 25,846,244 & 26,049,938 \\
    
    Small Model Output Tokens
    & 2,986,414 & 2,886,570 & 2,828,277 & 2,823,600 \\
    
    Large Model Input Tokens
    & 3,582,041 & 3,660,392 & 3,411,146 & 2,652,560 \\
    
    Large Model Output Tokens
    & 228,614 & 239,339 & 229,266 & 196,881 \\

    Large Model Calls 
    & 1,187 & 1,193 & 1,140 & 1,141 \\

    \midrule
    
    Small Model price \$ 
    & 8.68 & 8.35 & 8.12 & 8.14 \\
    
    Large Model price \$ 
    & 7.62 & 7.80 & 7.28 & 5.70 \\
    
    Total price \$ 
    & 16.30 & 16.15 & 15.40 & 13.84 \\
    \bottomrule
\end{tabular}
\caption{ALFWorld with \qwensmall + \qwenlarge. Token usage and cost comparison across different uncertainty metrics.}
\label{tab:alfworld_detailed_tokens_qwen_qwen}
\end{table}
  \begin{table}[ht]
\centering
\begin{tabular}{l|rrrr}
    \toprule
    & MTE & SP & PPL & Random \\
    \midrule
    Small Model Input Tokens
    & 25,724,005 & 25,400,456 & 24,894,956 & 26,120,373 \\

    Small Model Output Tokens
    & 1,381,591 & 1,382,268 & 1,367,765 & 1,458,582 \\
    
    Large Model Input Tokens
    & 3,827,339 & 3,554,058 & 3,833,118 & 2,480,882 \\
    
    Large Model Output Tokens
    & 72,139 & 66,605 & 67,572 & 47,121 \\

    Large Model Calls 
    & 1,446 & 1,349 & 1,426 & 1,099 \\

    \midrule
    
    Small Model price \$ 
    & 23.85 & 23.57 & 23.11 & 24.27 \\
    
    Large Model price \$ 
    & 7.71 & 7.15 & 7.65 & 5.00 \\
    
    Total price \$ 
    & 31.56 & 30.72 & 30.76 & 29.27 \\
    \bottomrule
\end{tabular}
\caption{ALFWorld with \llamathree + \gpt. Token usage and cost comparison across different uncertainty metrics.}
\label{tab:alfworld_detailed_tokens_llama_gpt}
\end{table}
  \begin{table}[ht]
\centering
\begin{tabular}{l|rrrr}
    \toprule
    & MTE & SP & PPL & Random \\
    \midrule
    Small Model Input Tokens
    & 26,235,704 & 25,074,274 & 25,778,042 & 27,914,399 \\
    
    Small Model Output Tokens
    & 1,613,201 & 1,543,982 & 1,596,770 & 1,679,948 \\
    
    Large Model Input Tokens
    & 3,994,820 & 4,108,710 & 4,278,414 & 2,692,680 \\
    
    Large Model Output Tokens
    & 275,371 & 271,067 & 282,763 & 197,577 \\
    
    Large Model Calls 
    & 1,445 & 1,383 & 1,481 & 1,139 \\

    \midrule
    
    Small Model price \$ 
    & 24.51 & 23.42 & 24.09 & 26.04 \\
    
    Large Model price \$ 
    & 8.54 & 8.76 & 9.12 & 5.78 \\
    
    Total price \$ 
    & 33.05 & 32.18 & 33.21 & 31.82 \\
    \bottomrule
\end{tabular}
\caption{ALFWorld with \llamathree + \qwenlarge. Token usage and cost comparison across different uncertainty metrics.}
\label{tab:alfworld_detailed_tokens_llama_qwen}
\end{table}

  \begin{figure}[h!]
    \centering
    \includegraphics[trim={0.cm 0.cm 0.cm 0.cm},clip,width=0.6\linewidth]{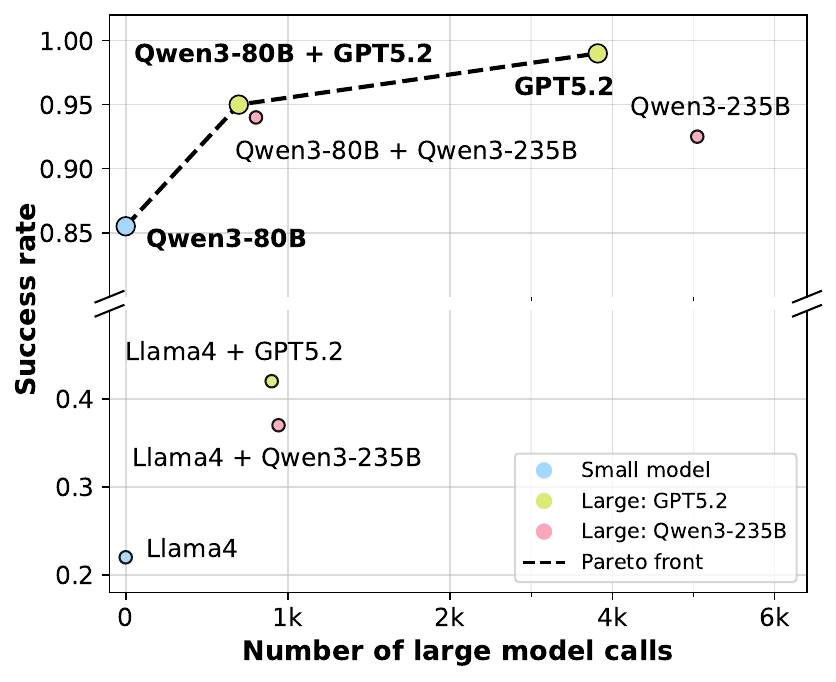}
    \caption{\textbf{Pareto front of success rate vs. large model calls on MiniGrid with perplexity-based deferral.} ReDAct enables small models to approach large model performance while using only ~15\% of large model calls.}
    \label{fig:pareto_minigrid_calls}
  \end{figure}
  
\end{document}